\documentclass[oneside, a4paper, onecolumn, 11pt]{article}
\usepackage[a4paper,left=2cm,top=2cm,bottom=1.5cm,right=2cm]{geometry}
\usepackage{times} 
\usepackage{amsmath}
\usepackage{tabularx}
\usepackage{makecell}
\usepackage{graphicx}
\usepackage[dvipsnames]{xcolor}

\begin{document}

\title{Model-Based and Data-Driven Strategies in Medical Image Computing}


\author{Daniel Rueckert\thanks{D. Rueckert is with the Department of Computing, Imperial College London, UK. E-mail: d.rueckert@imperial.ac.uk}\; and Julia A. Schnabel\thanks{J. A. Schnabel is with the Department of Biomedical Engineering, King's College London, UK. E-mail: julia.schnabel@kcl.ac.uk}
}

%

\maketitle

\begin{abstract}
Model-based approaches for image reconstruction, analysis and interpretation have made significant progress over the last decades. Many of these approaches are based on either mathematical, physical or biological models. A challenge for these approaches is the modelling of the underlying processes (e.g. the physics of image acquisition or the patho-physiology of a disease) with appropriate levels of detail and realism. With the availability of large amounts of imaging data and machine learning (in particular deep learning) techniques, data-driven approaches have become more widespread for use in different tasks in reconstruction, analysis and interpretation. These approaches learn statistical models directly from labelled or unlabeled image data and have been shown to be very powerful for extracting clinically useful information from medical imaging. While these data-driven approaches often outperform traditional model-based approaches, their clinical deployment often poses challenges in terms of robustness, generalization ability and interpretability. In this article, we discuss what developments have motivated the shift from model-based approaches towards data-driven strategies and what potential problems are associated with the move towards purely data-driven approaches, in particular deep learning. We also discuss some of the open challenges for data-driven approaches, e.g. generalization to new unseen data (e.g. transfer learning), robustness to adversarial attacks and interpretability. Finally, we conclude with a discussion on how these approaches may lead to the development of more closely coupled imaging pipelines that are optimized in an end-to-end fashion.
\end{abstract}



\section{Introduction}

Medical 
imaging is playing a key role in many clinical applications, ranging from the detection and diagnosis of disease to the planning and monitoring of therapy as well as the guidance of interventions and surgery. Over the last decades, our ability to image the anatomy and function with ever greater spatial (and temporal) resolution has significantly improved. This has led to an increasing need to automatically extract quantitative information from medical images and to analyse and interpret this information. This is critical to support diagnostic and treatment approaches that are appropriately customized to each individual patient, leading to so-called personalized or individualized medicine \cite{Topol2014}. Similarly, the ability to extract quantitative information from medical images is crucial in supporting the efficient analysis of large-scale population studies such as UK Biobank \cite{Miller2016,Petersen2013} or the German National Cohort \cite{GNC2014}. These large scale population studies image 10,000's of subjects and offer the potential to identify new and tailored strategies for early detection, prediction, and primary prevention of major diseases. However, the traditional analysis of medical images via visual interpretation by human experts is not feasible for such studies.

To support the quantification of clinically useful information from medical images, several steps are necessary. One of the first steps along the imaging pipeline is the acquisition and reconstruction of images. For example, in magnetic resonance (MR) imaging the scanner acquires data in the k-space domain, and an image must then be reconstructed from the acquired k-space samples. Similarly, in X-ray computed tomography (CT) imaging, the acquired projection data must be reconstructed into an image by inverting the Radon transformation. Once images have been reconstructed from the acquired data, the images are often enhanced before further analysis is carried out. This image enhancement can include a denoising of the image in order to improve its signal-to-noise ratio, or the application of image super-resolution techniques in order to boost its spatial (or temporal) resolution. Another form of image enhancement is the registration and fusion of different images (either from different image sequences or different modalities altogether). This enables the integration of structural and functional information. After this, semantic image interpretation is used to answer the following question: What is in this image, and where in the image is it located? In other words, the aim is to locate and segment anatomical structures in the image (e.g. organs). In some applications it is also of interest to track anatomical structures across time, e.g. the contraction and relaxation of the heart in a cine sequence of cardiac MR images. Once the semantic image interpretation has been completed, it is possible to extract and quantify imaging biomarkers. These imaging biomarkers can be regarded as a measurable indicator of some biological state or condition. Biomarkers are therefore a prerequisite to examine normal biological processes (e.g. growth or aging), pathogenic processes, or pharmacological responses to a therapeutic intervention. For example, the volume of a tumor or its shape and texture can be used to characterise the tumor as benign or malignant and to assess whether the tumor is responding to radio- or chemotherapy. In disease diagnosis or prediction, it is important to identify anomalies which may correspond to disease-related pathologies. In an ideal scenario, one would like to predict diseases as early as possible before their onset. This increases the possibility of early treatment and therefore improve outcome for patients. 

As mentioned above, the extraction of clinically useful information from medical images is critically dependent on the ability to acquire and analyse medical images, ranging from reconstruction, enhancement, registration, localisation, segmentation and tracking as well as shape and appearance modelling. Many of these medical image computing tasks involve solving an ill-posed problem, e.g. reconstructing or denoising an image. To address these ill-posed problems, traditional approaches are often model-based in order to enable the regularization of the ill-posed problem by incorporating prior knowledge. Often these models are either geometric, physical or biological models. More recently, data-driven approaches based on machine learning, in particular deep learning, have revolutionized medical imaging, offering superior performance to traditional model-based approaches in many medical imaging tasks. These approaches often do not use an explicit geometric or physical model. Instead, they accumulate information from a very large number of labelled or unlabelled examples into a statistical model that is then used to perform a prediction task such as regression or classification. 

The aim of this article is to provide an assessment of the dichotomy between traditional model and data-driven approaches in medical imaging. Our objective is to highlight key trends from the recent past and present, and predict future trends. We start by outlining some of the most commonly used model-based approaches that have been traditionally deployed in medical imaging. We then compare and contrast these to more recent data-driven approaches which use less domain specific knowledge than their model-based counterparts but often outperform them. We discuss both approaches in terms of performance (robustness, accuracy and speed) as well as in terms of clinical utility (interpretability and explainability). We also examine some of the main challenges for purely data-driven machine learning approaches. We then outline how model-based and data-driven approaches are likely to converge in the future to address some of these challenges. Finally, we discuss the implications of this for medical imaging, in particular how these approaches may lead to the development of more closely coupled imaging pipelines that are optimized in an end-to-end fashion, offering the potential to revolutionize the field of medical imaging.

\section{Methods}

All steps of the medical imaging pipeline typically make extensive use of models. In this paper we define the term model in a very general fashion, in the sense that it provides a transformation of input (data) into the desired output. For example, in image reconstruction the model helps to transform the acquired sensor data into an image, in image registration the model is used to relate two images (inputs) via a transformation (output), and in image segmentation the model is used to transform an image in which each pixel corresponds to an intensity value, into an semantic segmentation where each pixel has a label corresponding to an organ, anatomical structure or pathology. In diagnosis, the model helps to map image derived features and biomarkers to a diagnostic label.

\subsection{Types of models}

In medicine and biology, models are often categorized into mechanistic models and phenomenological (statistical) models \cite{Baker2018}. In mechanistic models, a hypothesized relationship between the input and output is specified in terms of a mathematical, physical or biological process typically governed by a small number of parameters. In contrast, a phenomenological model seeks to use a generic statistical model to discover a hypothesized relationship between the input and output. In other words, the aim is to seek statistical relationships and correlations between inputs and outputs.

The list below describes some of the most commonly used models in medical imaging:
\begin{itemize}
    \item Generic models
    \begin{itemize}
        \item Mathematical models
        \item Biological or physics-based models
    \end{itemize}
    \item Probabilistic models
    \begin{itemize}
        \item Gaussian mixture models
        \item Graphical models, including Markov Random Fields (MRF) and Conditional Random Fields (CRF)
    \end{itemize}
    \item Population-based models
    \begin{itemize}
        \item Single-subject atlases
        \item Probabilistic atlases
        \item Statistical atlases (shape and appearance models)
    \end{itemize}
    \item Shallow learning models
    \begin{itemize}
        \item Regression
        \item Nearest-neighbor methods
        \item Support Vector Machines (SVM)
        \item Random Forests
    \end{itemize}
    \item Deep learning models
    \begin{itemize}
        \item Recurrent neural networks
        \item Convolutional neural networks
        \item Autoencoders
        \item Deep reinforcement learning
    \end{itemize}
\end{itemize}

For the purposes of this article, we refer to generic models, probabilistic models and population-based models as traditional model-based approaches whereas learning-based models (shallow or deep) are referred to as data-driven approaches.

\subsection{Model fitting}

In order to apply models to medical images, we typically fit our model to the data. This is traditionally achieved by solving an optimization problem where one minimizes a loss function (also called cost function or energy). The function usually consists of 
\begin{equation}
    {\cal D}(M(x, \phi), y) + {\cal R}(\phi)
\label{eq:model}
\end{equation}
The first term ${\cal D}$ is typically referred to as data fidelity term and measures how well the model instantiated by the parameters $\phi$ and the unobserved data $x$ explains the observed data $y$ (e.g. sensor or imaging data). The second term ${\cal R}$ is a regularization term that expresses prior knowledge with respect to the parameters of the model.

In traditional model-based approaches, the model fitting is performed individually for each image via optimization of the model parameters. In contrast to this, in machine learning approaches the model is optimized during the training stage for all images in the training set. When the machine learning model is then applied to new images (testing stage or inference), the model can be simply evaluated without any need for further optimization. As a result, machine learning models are slow during training but extremely fast during their application.

In the following, we briefly review different model-based approaches for a range of medical image computing tasks. Many of these approaches can be characterized based on the types of models that they employ for the respective tasks.

\subsection{Image Reconstruction}

Image reconstruction is a crucial step in MR, CT, positron emission tomography (PET), single-photon emission tomography (SPECT), and ultrasound imaging. It enables the transformation of information acquired by the imaging sensor from the sensor domain into the image domain so that it can be visually interpreted. In many cases, the transformation from the image domain to the sensor domain can be characterized by a well-understood mathematical forward model:
\begin{equation}
y = A(x) + \epsilon
\end{equation}
Here $x$ represents the image one would like to recover, $A$ represents the transformation from the image domain to the sensor domain, $y$ represents the sensor measurements and $\epsilon$ represents the measurement noise. The goal of image reconstruction is the recovery of the image $x$ from the measured sensor data $y$. For CT, $A$ can be largely modelled as a Radon transformation, while for MR imaging, $A$ is given by a Fourier transformation. During the image reconstruction, the forward model $A$ must be inverted, which can be challenging in cases where the sensor data is undersampled, affected by noise or sensor imperfections. In this case an analytic solution to the inverse problem may not exist.

For example, in MR imaging it is often desirable to use acquisitions that undersample in the sensor domain (the so-called k-space that represents the Fourier-encoded version of the image in the frequency domain) in order to acquire the image faster. In this case, the image can no longer be fully recovered by applying the inverse Fourier transformation.

Assuming that the noise is normally distributed, a common approach is to recover $x$ by solving the following least squares problem:
\begin{equation}
x = \operatorname*{arg\,min}_x \frac{1}{2}||A(x)-y||^2 
\end{equation}
Here the sums-of-squared difference measures the difference between the forward model and the measured data (and thus is equivalent to the data fidelity term ${\cal D}$ in equation \ref{eq:model}). However, solving for $x$ is usually an ill-posed problem, so it is common to add a regularisation term ${\cal R}$ to the problem:
\begin{equation}
x = \operatorname*{arg\,min}_x \left( \frac{1}{2}||A(x)-y||^2 + \lambda {\cal R}(x) \right)
\end{equation}
A common choice for the regularization term is the total variation norm of the reconstructed image, i.e. ${\cal R}=||\nabla x||_1$. This enforces a sparse set of gradients in the reconstructed image as the regularization term favours images with piecewise constant image intensities. However, this regularization does not always capture the intensity variations that naturally occur in medical images, so other regularization models are also often used. 

One very successful class of image reconstruction approaches are based on the theory of compressed sensing (CS) which provides strong theoretical guarantees in regards to the recovery of image $x$ \cite{Donoho2006,Lustig2007}. In these CS approaches, the model used for regularization is typically based on the assumption that the image to be recovered is sparse in some domain. This holds for images which can be compressed with little or no loss of perceptual image quality. Note that the assumption of sparsity does not need to hold directly for the image itself (even though some images may be naturally sparse, e.g. MR angiography) but for the image after a suitably chosen transform of the image into some other domain. This transform of the image is often chosen to be a wavelet or discrete cosine transform or the TV norm operator \cite{Lustig2007}. Another alternative is to learn adaptively the sparsifying transformation in order to optimally exploit the redundancy in the data \cite{Ravishankar2011}, e.g. by using dictionaries of local image patches.  

In all of the above approaches, $A$ represents the forward model of the imaging process and thus encodes the prior knowledge about the image reconstruction problem to be solved, and $R$ encodes the prior knowledge about the expected solution. For example, in the case of undersampled, single-coil MR images, the forward model can be described as $A(x) = {\cal S}{\cal F}x$ where ${\cal F}$ denotes the Fourier transformation and ${\cal S}$ denotes the undersampling mask. During the optimization, the forward model is used in order to evaluate how well the model explains the observed data.

\begin{figure*}
\centering
\includegraphics[width=0.95\textwidth]{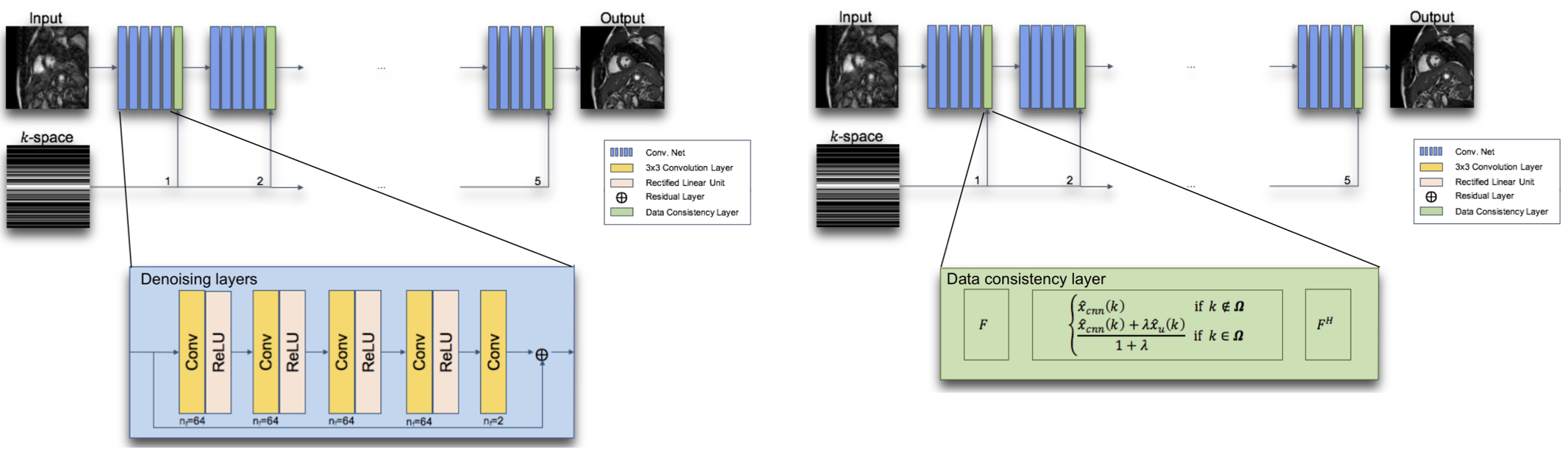}
\caption{\label{fig:recon}Example of a deep neural network for the reconstruction of dynamic MR images from undersampled k-space data \cite{Schlemper2018}. Here the deep neural network consists of a cascade of denoising layers (shown on the left, operating in the image domain) and data consistency layers (shown on the right, operating in the k-space domain).}
\end{figure*}

More data-driven approaches relax the assumption that the model used for regularization (e.g. total variation) is rather generic and thus cannot be dynamically adapted to the application. Instead, more data-driven approaches can learn the regularization model. For example, instead of assuming that the reconstructed image can be characterized by a piecewise constant function, one can learn a data-driven model $\Phi$ (e.g. a convolutional neural network \cite{LeCun1998}) of how the reconstructed image is likely to look \cite{Wang2016}:
\begin{equation}
x = \operatorname*{arg\,min}_x \left( \frac{1}{2}||A(x)-y||^2_2 + \lambda (||\Phi(A^*(y)) - x||^2_2) \right)
\end{equation}
Here $A^*$ is the adjoint operator. For example, in MR image reconstruction $A^*(x)$ yields the zero-filled reconstruction of the undersampled k-space, i.e. a reconstruction in form of an aliased image. An alternative to learning the regularization is to learn the iterative optimization, i.e. model fitting, directly \cite{Hammernik2018,Schlemper2018}. An example of a deep neural network for the reconstruction of dynamic MR images \cite{Schlemper2018} is shown in Figure \ref{fig:recon}. Here the iterative optimization process is unrolled into a deep cascade neural network.

Another strategy is to learn the mapping between the sensor domain and the resulting images through an entirely data-driven approach. One such as approach is AUTOMAP \cite{Zhu2017} which aims directly to estimate the transformation from the sensor domain to the image domain without any explicit knowledge of the mapping between the two domains. This is achieved by formulating the reconstruction problem as a supervised learning problem where the mapping is learned entirely using training data.

\subsection{Image Registration}

Image registration typically follows image reconstruction and addresses the problem of patient motion or repositioning between different scans. Registration is often used to fuse images of the same patient acquired from different imaging systems, such as MR and CT. This enables the integration and analysis of complementary information (e.g. structural and functional information). Image registration is also widely used to detect subtle structural or functional, disease or therapy-induced changes in serial imaging of the same patient over time. Another important application of registration is image guidance during surgical interventions, which requires mapping of the pre-operative to the intra-operative scene to enhance the information available during interventions. Finally, image registration is also a popular exploratory tool for comparing the anatomy or function between cohorts of patients or volunteers\cite{Ashburner2000,GASER20011140}, to further the understanding of disease processes across a population \cite{HUA2008458}, or to obtain more generic models of normal or abnormal development \cite{DBLP:journals/neuroimage/HuizingaPVRBIRN18}.

Registration describes the process of geometric, anatomical, metabolic or other functional alignment of two or more images, achieved via an optimization process that minimizes a cost function describing a data dissimilarity between the images, based on their intensities, extracted features or distances of anatomical landmarks or other markers; and a regularization cost that enforces the transformation to be well-behaved, thus avoiding physiologically inconsistent mappings from one image to another. Putting this into a model fitting context, the transformation $T$ describes the model to be fitted to the data (the images) using a registration cost function. The process of image registration is then defined as an optimization of transformation parameters $\phi$ with respect to the cost function, aiming to align an image $x$ to a reference image $y$:
\begin{equation}
\widetilde{\phi} = \operatorname*{arg\,min}_{\phi} \left( {\cal D}(T(x, \phi), y) + \lambda {\cal R}(\phi)) \right) \ ,
\end{equation}
Here the data fidelity term ${\cal D}$ describes the dissimilarity between the observed image $y$ and the transformed version of image $x$, viz. $T(x, \phi)$, and the regularization term ${\cal R}$ of the mapping parameters $\phi$ allows to incorporate prior knowledge about the smoothness of the transformation or other desirable properties of the mapping. In the following, we will link these traditionally model-driven registration concepts to more recent advances in data-driven strategies for image registration.

There is a significant amount of work on developing and testing suitable versions of image similarity terms defining ${\cal D}$ for intensity-based registration, ranging from sum of squared differences (SSD) or cross-correlation for single-modality registration, over mutual information \cite{DBLP:journals/mia/WellsVANK96,DBLP:journals/tmi/MaesCVMS97} and its normalized version \cite{DBLP:journals/pr/StudholmeHH99} specifically developed for multi-modality registration applications, to more generic, modality-independent measures such as proposed in  \cite{DBLP:journals/mia/HeinrichJBMGBS12}. In the simplest case of SSD, akin to the data consistency term commonly used for image reconstruction, and using no regularization (e.g. for a rigid body transformation), the optimization can be formulated as:
\begin{equation}
\widetilde{\phi} = \operatorname*{arg\,min}_{\phi}  \frac{1}{2}||T(x, \phi)-y||^2_2 
\end{equation}
The similarity measures described above make specific assumptions about the relationship between image intensities from the relevant scanning systems, including noise models, local or global intensity distributions and presence of intensity-derived features, inherently forming a prior to be used for the registration task at hand. An alternative approach is to infer a measure of image similarity directly from the images, using a larger, pre-registered image database. This is often referred to as {\it metric learning}. An example for similarity metric learning is presented in \cite{DBLP:conf/cvpr/BronsteinBMP10}, which developed a method for cross-modality metric learning using a similarity-sensitive hashing method, starting from a database of perfectly aligned T1/T2 weighted MR image pairs, and then learning from positive and negative pairs of local image patches. Such an approach is only possible if registered data are available. In the T1/T2 MR case, these data are inherently registered, rendering the need for registration somewhat obsolete; but for MR/CT cases, perfect registrations are not normally available, making it difficult to provide the necessary training data.

Another important aspect of image registration is the transformation model, which traditionally contains a strong prior for the expected types of motion. For example, a rigid-body transformation may be described by six degrees of freedom, $\{t_x, t_y, t_z, \alpha, \beta, \gamma\}$, i.e. three translations and three rotation angles. For more complex motions or deformations, compact parametric transformation models such as free-form deformations using B-splines \cite{DBLP:journals/tmi/RueckertSHHLH99} have become very popular due to their easy manipulation and inherent smoothness properties which are desirable in many, though not all medical imaging applications. Alternatively, deformations can be densely defined at the pixel or voxel level, ranging from optical-flow type methods based on the seminal work of \cite{DBLP:journals/ai/HornS81}, to the family of demons using stationary velocity fields and enforcing diffeomorphic mappings that avoid folding of the deformation field \cite{DBLP:journals/neuroimage/VercauterenPPA09}, with many variants existing, such as enforcing incompressibility of tissue deformation during motion tracking \cite{DBLP:journals/ijcv/MansiPSDA11}, symmetry of deformations \cite{DBLP:journals/neuroimage/LorenziAFP13} or allowing for very large deformations without causing tissue tearing \cite{DBLP:journals/ijcv/LombaertGPAC14}. Other prominent methods include Large Diffeomorphic Distance Metric Mapping (LDDMM) \cite{DBLP:journals/ijcv/BegMTY05}, DARTEL \cite{DBLP:journals/neuroimage/Ashburner07} or SyN \cite{DBLP:journals/mia/AvantsEGG08}. All these methods either have built-in regularization constraints offered by the mathematical formulation of the optimization, or  directly regularize the deformation field or its updates. Such explicit regularizers range from applying the Laplacian to the deformation parameters $\phi$, i.e. ${\cal R}=\nabla^2 \phi$, to more complex, locally adaptive regularization methods such as bilateral filters that allow to model locally discontinuous motion, which is relevant at sliding organ interfaces such as the lung surfaces \cite{DBLP:journals/mia/PapiezHFRS14}.

Recently, more data-driven techniques based on deep learning for image registration have emerged that combine the concepts of the traditional registration cost function, with learning and predicting the transformation parameters using a convolutional neural network (CNN) approach. These are commonly starting off from reformulating the traditional registration cost function as a loss function that needs to be minimized as part of the CNN training:
\begin{equation}
\widetilde{\phi} = \operatorname*{arg\,min}_{\phi} {\cal L}(x, y, \phi)
\end{equation}
where the loss function ${\cal L}$ of a CNN can be formulated as 
\begin{equation}
{\cal L}(x, y, \phi) = {\cal D}(T(x, \phi), y) + \lambda {\cal R}(\phi)  
\end{equation}
featuring again a data fidelity term ${\cal D}$ and a regularization term ${\cal R}$ operating on the image pair $(x, y)$ and the transformation parameters $\phi$. Using the basic concept of a loss function, deep learning registration approaches can be categorized either as {\it supervised}, starting from a training set of registered images whose transformation parameters are known, or as {\it unsupervised}, directly learning the transformation parameters from unregistered pairs of images.

Early examples using fully data-driven or deep-learning approaches include optical flow estimation approaches such as {\it DeepFlow} \cite{DBLP:conf/iccv/WeinzaepfelRHS13} or {\it FlowNet} \cite{DBLP:conf/iccv/DosovitskiyFIHH15}. While {\it DeepFlow} presents a matching algorithm that builds upon a multi-stage architecture akin to CNNs using an explicit variational framework, {\it FlowNet} and its derivatives are directly based on (stacked) CNN architectures. For example, two sequentially adjacent input images can be stacked together and fed through the network to learn their spatial relationships. {\it SpyNet} \cite{DBLP:conf/cvpr/RanjanB17} is a pyramidal, coarse-to-fine and more lightweight variant of {\it FlowNet}, with fewer hyper-parameters needed for optimization. Broadly following these optical flow approaches, the {\it Quicksilver} method  \cite{DBLP:journals/neuroimage/YangKSN17} is based on LDDMM \cite{DBLP:journals/ijcv/BegMTY05}, involving a deep regression model for predicting transformation parameters $\phi$ using image appearances, starting from LDDMM estimations for image patches as an initialisation.  

Unsupervised approaches that do not rely on pre-registered training data include the {\it Deep Learning Image Registration} (DLIR) framework \cite{DBLP:journals/mia/VosBVSSI19}, applied to cardiac and chest imaging, and the {\it VoxelMorph} method \cite{DBLP:conf/cvpr/BalakrishnanZSG18} and its diffeomorphic extension \cite{DBLP:conf/miccai/DalcaBGS18}, applied to brain MR imaging. Similar approaches have been proposed for cardiac motion tracking. The method proposed in \cite{DBLP:conf/miccai/QinBSPPNR18} performs cardiac image segmentation and motion tracking simultaneously while \cite{DBLP:journals/corr/abs-1907-13524} proposes a CNN-based model  to derive a probabilistic motion model from sequences of cardiac MR images.

For many of these approaches, a key component is the concept of spatial transformer networks presented in \cite{DBLP:conf/nips/JaderbergSZK15}. Spatial transformers are a neural network module that can be inserted into other neural network architectures, as a means to spatially transform feature maps. Most importantly, they allow for a large number of classes of transformations. Inserted as a spatial transform layer to the registration network, unsupervised end-to-end learning is accomplished, achieving similar accuracy in intersubject brain MR registration to traditional image registration methods such as those benchmarked in \cite{DBLP:journals/neuroimage/KleinGAYFAGMP10}, but with significant speedup. In \cite{DBLP:conf/miccai/WrightKGSMHRS18}, the concept of spatial transformers was successfully applied as a co-transformer network to simultaneously align fetal MR and ultrasound images to a common space. Finally, adversarial attacks using generative adversarial networks (GAN) \cite{DBLP:conf/nips/GoodfellowPMXWOCB14} have been exploited in \cite{DBLP:conf/miccai/FanCXYS18} to develop a framework which connects a registration network and a discrimination network with a deformable transformation layer, using feedback from the discrimination network in lieu of an explicit similarity measure.

The data-driven approaches for medical image registration share many commonalities with model-based approaches, in that they tend to have a cost (or loss) function, with a data fidelity (similarity) term and some form of regularization. Training of registration networks commonly involves iterations akin to traditional registration optimization, which can extend to one-shot registration at run-time after training. Training data typically still need to reflect the range of expected transformations and image modalities, as well as the variability of anatomy and patho-physiology. While deep learning approaches start to approach, and in some cases even exceed, the accuracy of conventional image registration, their key advantage is their computational speed, paving the way for on-line registration and subsequent, potentially integrated image analysis tasks.

\subsection{Image Segmentation}

Semantic image segmentation is a crucial task in many medical imaging applications. In this domain, nearly all successful approaches have been model-based. This is due to the fact that image segmentation is often a challenging problem complicated by image-related noise and artefacts as well as morphological variability of the anatomy. This means that the semantic segmentation of an image is an ill-posed problem, and models are used to regularize the segmentation and to encode prior knowledge about the intensity distribution of anatomical regions as well as their shapes. One of the earliest models used for medical image segmentation are so-called deformable models \cite{McInerney1996} which impose generic smoothness priors on the shape of the objects (or organs) of interest. These deformable models use ideas from geometry (e.g. splines) to efficiently represent organ shape while the shape constraints employed as well as the model fitting process are inspired by physics. 

In contrast to approaches that model the shape of organs using geometry- or physics-based models, probabilistic models aim to model the intensity distribution within organs and in the surrounding structures. One of the most successful approaches for modelling intensity distributions is based on Gaussian Mixture Models (GMM). One of the earliest applications of GMMs uses prior knowledge of intensity distributions and bias field (as caused by the B0 inhomogeneity of the MR scanner) to segment brain MR images \cite{Wells96}. Here the GMM is fitted to the image data using the Expectation Maximization (EM) algorithm. While this GMM uses prior knowledge about the intensity distribution, it does not use any prior information about the shape of the organs of interest. In \cite{Leemput99}, a probabilistic atlas is registered to the image to be segmented and defines a voxel-wise prior probability about the organs or tissue or interest. Additionally, a Markov Random Field (MRF) is used to further regularize the segmentation in the presence of image noise. In addition, it is possible to jointly carry out registration and segmentation in a unified probabilistic framework \cite{Ashburner2005}.

A probabilistic atlas typically encodes information about the average shape of organs/tissues across a large population. However, during the construction of the probabilistic atlas, details of the variations of the shapes are lost due to the averaging of shape information. Instead of using probabilistic atlases, Cootes et al. \cite{Cootes95} proposed the idea of active shape models that learn the natural shape variation via a statistical shape model. This statistical shape model is derived from a set of aligned training shapes which are analyzed using techniques such as principal component analysis (PCA) to build a parameterized, low-dimensional representation of the shape variation in the training set. As the statistical shape model is derived from a set of example or training shapes, this approach can be viewed as an early machine learning-based approach. However, in contrast to more recent machine learning-based approaches which perform segmentation as a classification or prediction task, the segmentation is here performed by fitting the learned shape model to image data via an iterative optimization approach. These statistical models can also be extended to learn not only shape but also appearance \cite{Cootes98}.

Instead of building statistical shape and appearance models, it is also possible to use the training shapes/images as individual atlases which are then combined into multi-atlas segmentation approaches \cite{Heckemann2006, Aljabar2009}. Here each individual atlas is registered to the target image using non-rigid image registration. The shapes or labels from the atlas can then be transferred onto the target image. However, each atlas registration is likely to introduce small segmentation errors that are caused by errors in the registration and in the annotations in the atlas. Assuming that these errors are uncorrelated, it is possible to combine these imperfect segmentations into a high-quality segmentation using ideas from classifier fusion \cite{Artaechevarria2009, Wang2013}.

More recently, data-driven approaches have dominated the area of medical image segmentation. As mentioned above, these approaches treat the segmentation task as a dense classification or prediction task. They do not use an explicit model to encode a-priori information about the shapes and intensity distributions. Instead, the a-priori information is implicitly encoded by examples that are  used during training. Early approaches have used support vector machines (SVM) \cite{Cortes1995} for segmentation, in particular for the segmentation of brain tumors \cite{10.1007/978-3-642-23626-6_44} or for the segmentation of the cartilage of the knee \cite{6116655} in multi-modal MR images. Other machine learning approaches that have been successfully used for segmentation include k-nearest neighbor classifiers, e.g. for brain tissue segmentation  \cite{VROOMAN200771,WARFIELD200043}. Many of these classification approaches use rather simple pixel- or voxel-wise intensity features to perform segmentation and thus often rely on the use of multi-spectral intensity information or anatomical priors such as probabilistic atlases for robust performance. Some of the most successful classical machine learning approaches for image segmentation are based on random or decision forests \cite{Criminisi2012}. In contrast to previous machine learning approaches, random forests allow the use of very large feature sets for each voxel (e.g. Haar-based features) as they perform feature selection and learning at the same time. Different variants of random forests have been successfully used for a wide range of tasks such as brain tumor segmentation \cite{Zikic2012} and abdominal organ segmentation \cite{Montillo2011}.

\begin{figure*}
\centering
\includegraphics[width=0.9\textwidth]{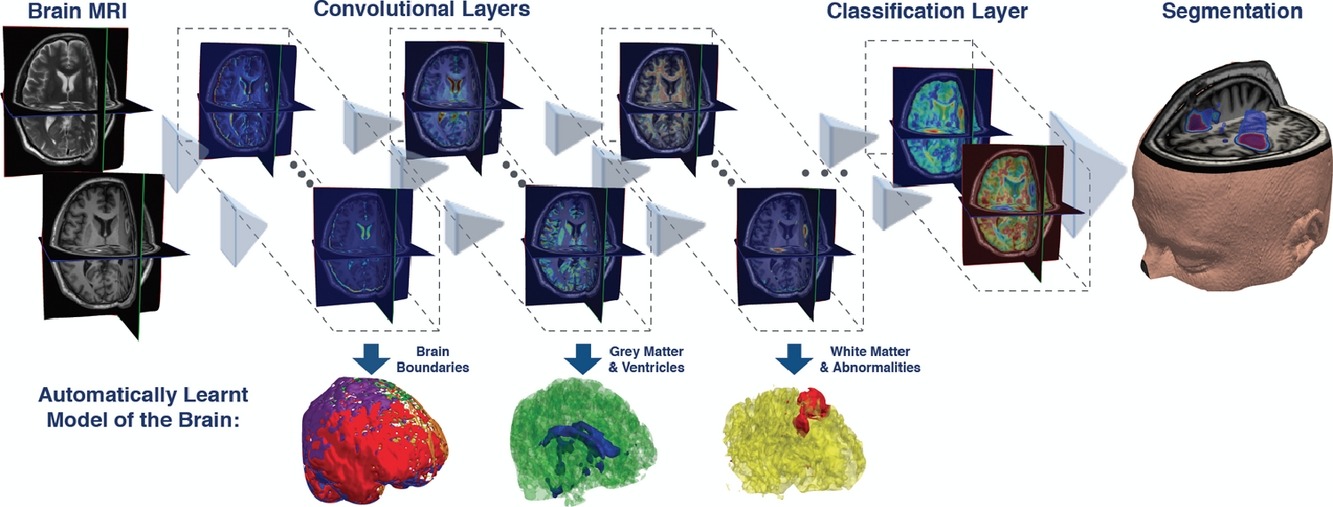}
\caption{\label{fig:deepmedic}Overview of the DeepMedic segmentation framework \cite{Kamnitsas2017}.}
\end{figure*}

However, the most successful segmentation approaches to date are based on deep neural networks, in particular convolutional neural networks (CNN). Several different CNN architectures have been proposed for this, including fully convolutional networks (FCN) \cite{Long2015}, the U-Net architecture \cite{Ronneberger2015} and the DeepMedic architecture \cite{Kamnitsas2017} (as shown in Figure \ref{fig:deepmedic}). A key ingredient for the success of CNNs is the fact that these approaches no longer require hand-engineered features (in contrast to classical machine learning approaches). Instead, the features are trainable and are learned directly from the training samples. There are numerous examples of the successful application of CNNs for semantic image segmentation, including for cardiac segmentation  \cite{Tran2016, Bai2018} as well as brain tumor or lesion segmentation \cite{Kamnitsas2017, Havaei2017}. These deep learning approaches generally outperform traditional model-based approaches not only in terms of segmentation accuracy but also in terms of computational speed: While the training of these models is computationally expensive, the application of the models to new images at inference time is usually very fast and efficient since no iterative model fitting is required. While most CNN-based approaches for semantic segmentation do not use any regularization in order to constrain the segmentation, approaches have been proposed that use conditional random fields (CRF) \cite{Kamnitsas2017} or learned shape priors \cite{Oktay2018} to constrain the segmentation.

\subsection{Image Interpretation}

Model-based approaches are frequently used for image classification, e.g. to identify whether a patient's anatomy contains pathologies or not. Many of these approaches model the geometry of the patient's organs and then use shape analysis techniques \cite{Bookstein1991} to perform classification. Often, these techniques analyze the geometry of the object surface by decomposing the shape into a set of basis function. One example of this is the work in \cite{Styner2004} which identifies shape variations in the hippocampus that are characteristic of schizophrenia. Alternatively, it is possible to analyze the organ shapes using a statistical shape model. These statistical shape models are similar to the model used for image segmentation \cite{Cootes95} and yield a parameterization of shape of the organ of interest in terms of the modes of variation \cite{Lindner2013} that can then be used for classification or prediction.

Atlas-based models can also be used for classification. These approaches compare individual subjects or groups to an population atlas, either in terms segmentation (e.g. voxel-based morphometry \cite{Ashburner2000}) or in terms of deformations required to deform the image into the atlas configuration (e.g. deformation- or tensor-based morphometry \cite{GASER20011140}). Both approaches has been used successfully for the diagnosis of neurodegenerative diseases such as Alzheimer's disease \cite{Kloeppel2008,Koikkalainen2011}.

More recently, machine-learning (in particular deep learning) approaches have dominated attempts to perform classification of images with the purpose of disease diagnosis \cite{Rajpurkar2018,Bien2018}. Many of these approaches use the image intensities directly without any explicit feature extraction. Instead, these approaches use CNNs, often with architectures similar to the VGG network \cite{Simonyan2014} that has been successfully used in many image classification tasks such as ImageNet. Such deep neural networks cannot only be used for disease classification, but also for detecting anatomical structures \cite{Vos2017}, identifying the correct scan plane in fetal ultrasound images \cite{Baumgartner2017}, detecting image artefacts in cardiac MR images \cite{Oksuz2018} or assessing the image quality in fetal ultrasound images \cite{Wu2017}. Finally, similar approaches can also be used for clinical decision support \cite{Fauw2018, Arbabshirani2018, Esteva2017} or for prediction of patient survival \cite{Bello2019}.

\section{Challenges and Directions for Future Work}

The use of model-based approaches in medical imaging has led to a number of breakthroughs over the last decades. More recently, the use of data-driven models has significantly increased as they outperform traditional model-based approaches both in terms of accuracy and speed. However, traditional model-based approaches often offer other advantages, e.g. in terms of generalisation, robustness, interpretation and validation. Some of the advantages and disadvantages of either type of approaches are outlined in Table \ref{tab:adv}. In the following we highlight some of the key requirements for the successful adoption of both traditional as well as data-driven models in clinical practice. We also discuss some of the open challenges and opportunities for future work in more detail.

\begin{table*}[htbp]
\centering \footnotesize
\begin{tabular}{|c|c|c|} \hline
    & Traditional models & Data-driven models  \\ \hline
Accuracy, Precision & \makecell{Medium, often limited by model\\ complexity and realism} & \makecell{Very high, but sometimes \\limited by training data}\\ \hline
Robustness & \makecell{High while model assumptions are valid, \\low if model assumptions violated} & \makecell{Low to medium, e.g. can be \\susceptible to adversarial attacks}
\\ \hline
Generalizability & \makecell{Extrapolation possible; \\ Usually generalize well} & \makecell{Generalization ability depends on representa- \\tiveness and variation within the training set}\\ \hline
Interpretability & \makecell{Small number of parameters; \\ Mechanistic relationship between \\ parameters and model behaviour}
& \makecell{Very large number of parameters; Black-box \\ relationship between parameters and model \\ behaviour makes interpretation challenging}
\\ \hline
Validation & Requires validation data & \makecell{Requires separate validation data \\ for each application domain} \\ \hline
Speed & \makecell{Slow as model fitting requires \\iterative optimization}& \makecell{Slow during training; \\Very fast during inference}
\\ \hline
Data & 
Requires no or limited amounts of annotated data
& 
Requires very large amounts of annotated data
\\ \hline
\end{tabular}
\vspace*{1.5ex}
\caption{\label{tab:adv}Advantages and disadvantages of traditional model-based and data-driven approaches}
\end{table*}

\subsection{Performance and Robustness}

Given a dataset with available ground truth, one can evaluate the performance of model-based approaches by computing the {\em accuracy} metric that is appropriate for the specific task (e.g. in the case of image reconstruction, the peak signal-to-noise ratio (PSNR), or in the case of image segmentation, the Dice metric). In machine learning, the accuracy of a model is also often referred to as model bias as it measures the difference between the model's prediction and the correct solution. However, the term bias can be misleading because it has different connotations in other areas, such as as medicine and law. 

A second useful metric is the model {\em precision}, which measures the variability of the model's prediction for a given input, assuming that a different dataset was used to optimize the model's parameters or hyper-parameters. Both of these quantities are useful to characterize the quality of models in general. For example, low accuracy indicates {\em underfitting}, i.e. the model is unable to accurately predict the desired output. This can indicate that the model is not sufficiently complex to capture the underlying patterns of the data. On the other hand a low precision indicates {\em overfitting}, i.e. the model is modelling the data as well as noise or overly complex data points.

In terms of accuracy, data-driven models often outperform traditional model-based approaches. For example, as we have seen earlier in image segmentation, nearly all state-of-the-art methods are based on data-driven deep learning techniques which often outperform other approaches by a significant margin. Whereas in traditional model-based approaches the model complexity is often limited, the model complexity of deep learning models is often significantly higher. However, a higher model complexity is also associated with a significantly large number of model parameters. Thus, deep learning models typically require very large training datasets in order to optimize the model parameters. At the same time, deep learning models with a large number of number of model parameters often overfit the training data, leading to poor performance away from the training data. This is in contrast to other model-based approaches that either require no training data or significantly fewer number of training examples.

Therefore, a common challenge for data-driven approaches is to build models that provide high performance but also high robustness. As mentioned before, the number of model parameters plays a crucial role here, but other factors also contribute to determining performance and robustness. For example, in deep learning models these factors include the number and type of layers (e.g. convolution, pooling, and fully-connected layers), the ordering of layers and the hyperparameters for each type of layer (e.g. the number of filters, the filter size, and stride). The number of possible choices makes the design space of deep neural networks architectures extremely large and difficult for exhaustive exploration. Thus, their design requires significant human expertise and computational effort. Recent approaches try to tackle this challenge through the automatic design of neural networks using techniques such as reinforcement learning \cite{DBLP:conf/iclr/BakerGNR17}.

\subsection{Generalisation ability}

Another important aspect of model-based approaches is their ability to generalise. In medical imaging the generalisation ability is typically defined in terms of how well the model can deal with unexpected variations in the input data. These variations can be caused by images acquired  across a wide range of scanners from different vendors and with different characteristics (e.g. different magnetic field strengths), different sequences (e.g. T1, T2, etc.), and different types of image noise (e.g. Gaussian, Rician or Poisson distributed noise) or artefacts (e.g. B0 inhomogeneity in MRI or streaking in CT). In addition, there is a significant variation across patients' anatomies across a population as well as the potential presence of pathology. Traditional model-based approaches can deal with such variations well if these are included in the model assumptions. However, the violation of the model assumptions can lead to poor generalization ability. Data-driven approaches have the advantage that no explicit assumption about noise, artefacts or anatomical or pathological variations have to modelled. However, they entirely depend of the training data being representative on the noise, artefacts and patient variability that can be encountered during deployment.


The key challenge of generalization is directly linked to the transfer of knowledge across multiple situations or domains. A good model should be able to generalize well and thus should be able to transfer knowledge across different domains. For traditional model-based approaches this is true as long as the model assumptions (or prior knowledge) are general enough to also be valid for each new domain. In contrast to traditional model-based approaches which explicitly encode the prior knowledge, data-driven approaches derive their prior knowledge purely from the training data (also called the source domain). It is often more difficult to assess how representative the training set is of the data that will be encountered in a new domain (also called the target domain). For example, a segmentation model might have been exclusively trained on data from the source domain, e.g. MR images acquired from MR scanners with a field strength of 1.5T, but the model is then deployed on a different target domain, e.g. images from MR scanners with a field strength of 3T. The shift in the distribution of the input data across the two domains is likely to cause a model to fail on the target domain. 

One strategy to deal with this challenge is to re-train the model from scratch for each new target domain. However, this requires a large amount of annotated data for the target domain and thus is often impractical. An alternative strategy is to use transfer learning which aims to use a model that has been pre-trained in the source domain in order to refine the model using small amounts of data from the target domain. In the arena of deep neural networks, it has been shown that such strategies outperform training neural networks from scratch \cite{Shin2016} and can be used to deal with variability across imaging sequences \cite{Ghafoorian2017}. A disadvantage of conventional transfer learning approaches is that they still require some annotated data from the target domain. Alternatively, it is possible to use adversarial learning approaches \cite{Goodfellow2015} in order to force deep neural networks to learn features that are domain invariant and thus generalize well to new domains \cite{Kamnitsas2017a,pmlr-v80-hoffman18a}. 

\subsection{Safety}


In addition to being accurate, robust and generalisable, model-based solution need to be resilient and safe. This leads to the challenge on how to formally verify the correctness of a model. While at least some traditional models can be verified in a formal setting (i.e. the correctness of the model can be proved or disproved with respect to a certain formal specification), this verification is much more challenging for data-driven models such as neural networks \cite{Leofante2018}, especially in a formal setting. Instead of using formal verification, recent approaches have focused on adversarial attacks on neural networks. These attacks work by attempting to modify the input in such a way that the output changes away from the desired output. In the context of images, this requires perturbations such as the addition of a "€œnoise-like"€ intensity pattern. Such adversarial attacks were first convincingly demonstrated in \cite{Goodfellow2015}. One remarkable aspect of these attacks is the fact that the perturbations that are required to be added to the image to successfully attack the model are often very small and visually hard to perceive. It has been demonstrated that this also applies to the medical imaging domain \cite{Finlayson2019}.  More recently, advances have been made in the area of providing guarantees about the performance and robustness of neural networks \cite{Katz2017}. Automated reasoning techniques have been proposed by several researchers in order to close the gap between neural networks and applications requiring formal guarantees about their behavior. A summary of existing approaches for the automated verification of neural networks can be found in \cite{Leofante2018}.

\subsection{Transparency}


Another important challenge is related to the transparency of model-based solutions to allow users to understand their capabilities as well as their limitations.  In traditional model-based approaches these  capabilities and limitations can often be derived from the underlying explict model. However, data-driven approaches often use implicit (or so-called {\em black-box}) models. These black-box models are defined by the mathematical model of the machine learning algorithm being used (e.g. a deep neural network). However, these are very complex mathematical models which often have a very large number of parameters, making it difficult to interpret these parameters or to predict the model's behaviour, i.e. when applied to unseen data. In addition to predicting the model's behaviour, it is often desirable to understand the model's behaviour, e.g. why the model fails when applied to one image but not on another image.

Most of the approaches that focus on making deep neural networks more interpretable, aim to provide some form of saliency map \cite{Simonyan2014a,Springenberg2015} that highlights which regions in the image are important for a classification decision. An alternative strategy is to use a so-called attention mechanism that automatically learns to focus on target structures of interest. This can also be used to improve the performance of deep neural networks in tasks such as classification and segmentation \cite{Schlemper2019}. Other approaches focus on visualizing the filters learned by deep neural networks \cite{Zeiler2014}, identifying important features \cite{Shrikumar2017} or reconstructing images from features learned by the deep neural network \cite{Mahendran2015}.

While interpretability of models is important from a technical point of view, the ability to explain the output of models is even more important in the context of certain applications. As a result, there is a growing interest in explainable machine learning models, even though the field is still in its infancy \cite{Gunning2019}. This importance also increases with the level of autonomy with which the models are deployed in a clinical settings. For example, one may use deep neural networks to predict the segmentation of a brain tumour in an MR image. This segmentation can be visually assessed by a radiologist to verify the accuracy of the output of the model. Here, the model is deployed in an assistive fashion, i.e. the radiologist can intervene and take corrective action. However, if one uses a model for the diagnosis or prediction of diseases, it becomes more difficult to judge the trustworthiness of the model-derived solution. In order for the user (i.e. the clinician) to be able to assess the trustworthiness, it becomes critical to be able to explain the output of the model. A particular challenge is that deep learning models, such as CNNs, often make overconfident predictions with poor generalization on unseen data \cite{DBLP:journals/corr/abs-1812-05720}. To account for this, recent approaches propose solutions that learn explicit uncertainty measures which capture the confidence of the system in the predicted output \cite{DBLP:journals/corr/abs-1906-07775}. Such measures of uncertainty can then be used for improving diagnostic performance, e.g. by referring cases with the most uncertain decisions for further inspection by a human expert \cite{Leibig2017}.
 

A related challenge is that of bias and fairness. For example, a data-driven model that predicts outcome or survival from medical images (e.g. \cite{Bello2019}) may be trained with data containing certain population characteristics (e.g. ethnicity or gender) and thus be only accurate if applied to patients with the same characteristics. However, a particular problem is that such a bias may often be hidden in the data that have been used to develop these models, and thus may be quite hard to detect. From a technical point of view, the bias can be addressed via transfer learning approaches that have been already mentioned earlier. Even with the bias correctly identified, it may be difficult to address the causes of the bias. For example, it may be logistically difficult or economically costly to acquire additional training data for the target domain. This raises the challenge of ensuring the fairness of these solutions and how this fairness can be measured and assessed. This is an active area of research, in particular in machine learning \cite{Dwork:2012:FTA:2090236.2090255,DBLP:journals/corr/Grgic-HlacaZGW17,DBLP:journals/corr/abs-1708-01870}.

\subsection{Data}

A key ingredient for data-driven approaches is the data available for training the machine learning models. In practice, the amount of data available during the development of the model is often limited. At the same time, the objective assessment of model accuracy requires the separation of the available dataset into training, validation and test sets in order to measure the model performance. The training set is used to determine the optimal parameters of the model. This is typically achieved by minimizing the loss function of the model on the training data. The validation dataset is then used to provide an unbiased estimate of the quality of the model that has been trained on the training dataset. By assessing the final model quality on the validation dataset, it is possible to tune the model's hyper-parameters. Finally, the optimized model must be evaluated on the test set (or holdout dataset) in order to obtain a final unbiased estimate of the model quality. Care has to be taken to avoid cross-contamination between the different splits of the dataset.

In many cases, the available data are split repeatedly into different training, validation and tests, enabling the use of cross-validation to not only estimate the accuracy but also the precision of the solution. A common approach uses nested cross-validation with an inner loop optimizing the model's hyper-parameters, while the outer loop evaluates the performance of the model with its optimal hyper-parameters. This allows for the estimation of the variance of the model quality. Furthermore, if the model is stable, its hyper-parameters should be the same across all runs of the outer loop. It is important to note that techniques such as cross-validation and bootstrapping can in some cases fail to produce accurate estimates of a model's accuracy \cite{Kohavi1995}. 

As mentioned already earlier, the amount of data available during model development is often limited. There are several reasons for this: One key factor is the amount of work required to annotate large amounts of medical imaging data. Since most current data-driven approaches rely on supervised machine learning approaches, the availability of annotated data is still a bottleneck for many applications. There are numerous approaches currently being developed to reduce the reliance of detailed annotations, e.g. by exploiting weakly labelled data \cite{DBLP:journals/tmi/RajchlLOKPBDRHK17,10.1007/978-3-319-66179-7_65}.

Another factor that limits the amount of data available for training models relates to the challenges associated with data sharing and privacy concerns. This is particularly important in the medical domain where images and associated clinical data contain highly sensitive and personal information. Once such data have left the hospital setting, the appropriate sharing and use of the data is very hard to control, leading to a reluctance of healthcare providers and patients to surrender control of these data. Technical solutions to this challenge that have recently attracted interest include approaches such as federated learning. Federated learning is a distributed machine learning approach which enables model training on a large corpus of decentralized data \cite{DBLP:journals/corr/abs-1902-01046}. A successful example of federated learning applied to medical image segmentation can be found in \cite{10.1007/978-3-030-11723-8_9}. Other approaches that use learning with differential privacy \cite{6582713} or learning with encrypted data \cite{pmlr-v48-gilad-bachrach16}, but to the best of our knowledge neither has been applied in the context of medical imaging.

\section{Discussion and Conclusion}

In this article we have reviewed model-based as well as data-driven strategies for medical image computing. Even though the distinction between model-based and data-driven strategies is somewhat arbitrary, both strategies differ in terms of how they define models. While traditional model-based approaches explicitly define mathematical, physical, statistical or probabilistic models that are used to solve a specific task, data-driven strategies such as deep learning often use complex black-box models with millions of trainable parameters. 

The introduction of deep learning models has revolutionized the field of medical image computing. In nearly all problem domains ranging from image reconstruction, segmentation and registration to interpretation, the current best performing state-of-the-art is based on deep neural networks. As mentioned in the previous section, the availability of data is is critical for the success deep learning approaches. Recently, several very large image databases without annotations (e.g. UK Biobank \cite{Miller2016, Petersen2013}) as well as large image databases with annotations (e.g. CheXpert \cite{Irvin2019} or DeepLesion \cite{Yan2018}) have become available.  At the same time, medical image databases with expert annotations are still at least one order of magnitude smaller than comparable databases in computer vision, e.g. ImageNet \cite{ILSVRC15} or MS-COCO \cite{10.1007/978-3-319-10602-1_48}. It seems clear that the availability of similar very-large scale medical image databases with expert annotations have the potential to boost the performance of deep learning approaches in medical imaging even further.
In addition to offering superior performance to traditional model-based approaches, deep learning approaches also offer significant speed-ups at run-time, making clinical deployment more realistic. A disadvantage of deep learning approaches is that they often do not generalize well beyond data that are very similar to the training data. In particular, the generalization ability of deep learning approaches is difficult to predict, often leading to failures during the clinical deployment.

At the same time, deep learning approaches are often very versatile, so that the same neural network architecture can be used for different purposes. For example, this means that a deep neural network with a same decoder/encoder architecture (e.g. U-Net \cite{Ronneberger2015, Ronneberger2016}) can be used for a variety of tasks such as image reconstruction, denoising and segmentation. This makes the development of prototypes for new medical imaging applications faster than traditional model-based approaches which often require a significant amount of hand-crafted engineering by the developer. In addition, deep learning models can be trained in an end-to-end fashion. This offers a number of advantages: For example, instead of defining image features that may be useful for image segmentation by hand, deep neural networks can learn the optimal feature representation at the same time as learning how to infer a segmentation from the features.

\begin{figure*}
\centering
\includegraphics[width=0.9\textwidth]{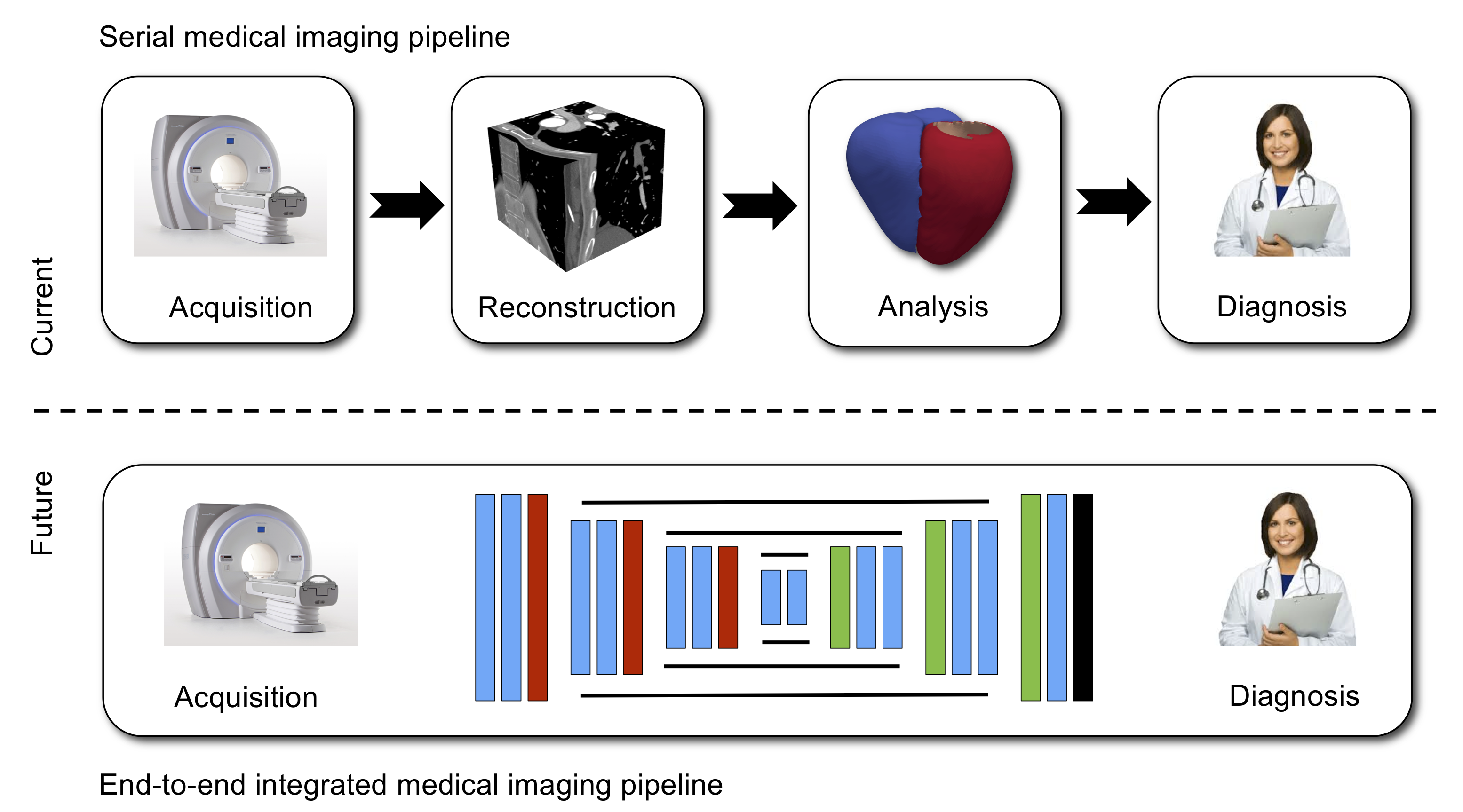}
\caption{\label{fig:future}In the current medical imaging paradigm, the process from image acquisition, reconstruction to analysis $\&$ interpretation and finally clinical diagnosis is an entirely serial process. A future paradigm of medical imaging offers the potential for a fully integrated, end-to-end optimized pipeline that is optimal for clinical decision making.}
\end{figure*}

It is also worthwhile to note that approaches that can be trained end-to-end may offer the potential in the future to change significantly the way medical imaging is performed. The current process of making clinical decisions based on medical imaging is essential sequentially (see Figure \ref{fig:future}, top): First, the sensor data is acquired by the image acquisition device (e.g. a MR scanner). Then an image is reconstructed from the sensor data. The resulting image is then analyzed, e.g. by performing image segmentation, in order to extract quantitative information for clinical decision making. This process is entirely serial with no feedback between the different stages of the medical imaging pipeline. This serial nature provides limited ability for adjustment of the upstream imaging pipeline based on downstream requirements and means that the different stages of imaging pipeline are not necessarily optimal for clinical decision making. For example, in a neurological imaging study, the image quality may not be sufficient for the clinical endpoint, e.g. assessing hippocampal atrophy. The end-to-end nature of deep learning approaches offers the potential to develop a future imaging pipeline where each stage of the imaging pipeline is closely coupled and can provide upstream and downstream feedback. This may enable the development of an integrated imaging pipeline that is optimized for the desired clinical endpoint (see Figure \ref{fig:future}, bottom) in an end-to-end fashion.

Finally, despite the recent successes of data-driven approaches, it is likely that the combination of data-driven and traditional model-based approaches is required to overcome some of the challenges that have been discussed in the previous section. In particular, with respect to challenges such as generalisability, explainability and data efficiency it is likely that the combination of traditional model- and data-driven approaches will lead to future advances. Several approaches have already demonstrated that priors in form of anatomical models can significantly improve the performance of applications such as segmentation \cite{8672808,DBLP:conf/ipmi/CloughOBSK19} or super-resolution reconstruction \cite{Oktay2018}.

\section*{Acknowledgment}

The authors acknowledge funding support from EPSRC (EP/P001009/1, EP/P023509/1, EP/N026993/1), the Wellcome Trust/EPSRC Centre of Medical Engineering (NS/A000049/1), the Wellcome Trust/EPSRC IEH Award (NS/A000025/1) and the Innovate UK London Medical Imaging and AI Centre for Value Based Healthcare.


\bibliographystyle{plain}
\bibliography{paper}

\begin{thebibliography}{100}

\bibitem{Aljabar2009}
P.~Aljabar, R.~A. Heckemann, A.~Hammers, J.~V. Hajnal, and D.~Rueckert.
\newblock Multi-atlas based segmentation of brain images: Atlas selection and
  its effect on accuracy.
\newblock {\em NeuroImage}, 46(3):726--738, 2009.

\bibitem{Arbabshirani2018}
M.~R. Arbabshirani, B.~K. Fornwalt, G.~Mongelluzzo, J.~Suever, B.~D. Geise,
  A.~A. Patel, and G.~J. Moore.
\newblock Advanced machine learning in action: identification of intracranial
  hemorrhage on computed tomography scans of the head with clinical workflow
  integration.
\newblock {\em npj Digital Medicine}, 1, 2018.

\bibitem{Artaechevarria2009}
X.~Artaechevarria, A.~Mu{\~{n}}oz{-}Barrutia, and C.~Ortiz{-}de{-}Solorzano.
\newblock Combination strategies in multi-atlas image segmentation: Application
  to brain {MR} data.
\newblock {\em {IEEE} Trans. Med. Imaging}, 28(8):1266--1277, 2009.

\bibitem{DBLP:journals/neuroimage/Ashburner07}
J.~Ashburner.
\newblock A fast diffeomorphic image registration algorithm.
\newblock {\em NeuroImage}, 38(1):95--113, 2007.

\bibitem{Ashburner2000}
J.~Ashburner and K.~J. Friston.
\newblock Voxel-based morphometry: The methods.
\newblock {\em NeuroImage}, 11(6):805 -- 821, 2000.

\bibitem{Ashburner2005}
J.~Ashburner and K.~J. Friston.
\newblock Unified segmentation.
\newblock {\em NeuroImage}, 26(3):839--851, 2005.

\bibitem{DBLP:journals/mia/AvantsEGG08}
B.~B. Avants, C.~L. Epstein, M.~Grossman, and J.~C. Gee.
\newblock Symmetric diffeomorphic image registration with cross-correlation:
  Evaluating automated labeling of elderly and neurodegenerative brain.
\newblock {\em Medical Image Analysis}, 12(1):26--41, 2008.

\bibitem{Bai2018}
W.~Bai, M.~Sinclair, G.~Tarroni, O.~Oktay, M.~Rajchl, G.~Vaillant, A.~Lee,
  N.~Aung, E.~Lukaschuk, M.~Sanghvi, F.~Zemrak, K.~Fung, J.~Miguel~Paiva,
  V.~Carapella, Y.~Jin~Kim, H.~Suzuki, B.~Kainz, P.~Matthews, S.~Petersen, and
  D.~Rueckert.
\newblock Automated cardiovascular magnetic resonance image analysis with fully
  convolutional networks.
\newblock {\em Journal of Cardiovascular Magnetic Resonance}, 20, 2018.

\bibitem{DBLP:conf/iclr/BakerGNR17}
B.~Baker, O.~Gupta, N.~Naik, and R.~Raskar.
\newblock Designing neural network architectures using reinforcement learning.
\newblock In {\em International Conference on Learning Representations (ICLR)},
  2017.

\bibitem{Baker2018}
R.~Baker, J.-M. Pe±a, J.~Jayamohan, and A.~Jerusalem.
\newblock Mechanistic models versus machine learning, a fight worth fighting
  for the biological community?
\newblock {\em Biology Letters}, 14:20170660, 2018.

\bibitem{DBLP:conf/cvpr/BalakrishnanZSG18}
G.~Balakrishnan, A.~Zhao, M.~R. Sabuncu, J.~V. Guttag, and A.~V. Dalca.
\newblock An unsupervised learning model for deformable medical image
  registration.
\newblock In {\em Computer Vision and Pattern Recognition (CVPR)}, pages
  9252--9260, 2018.

\bibitem{10.1007/978-3-642-23626-6_44}
S.~Bauer, L.-P. Nolte, and M.~Reyes.
\newblock Fully automatic segmentation of brain tumor images using support
  vector machine classification in combination with hierarchical conditional
  random field regularization.
\newblock In {\em Medical Image Computing and Computer-Assisted Intervention
  (MICCAI)}, pages 354--361, 2011.

\bibitem{Baumgartner2017}
C.~F. Baumgartner, K.~Kamnitsas, J.~Matthew, T.~P. Fletcher, S.~Smith, L.~M.
  Koch, B.~Kainz, and D.~Rueckert.
\newblock Sononet: Real-time detection and localisation of fetal standard scan
  planes in freehand ultrasound.
\newblock {\em {IEEE} Trans. Med. Imaging}, 36(11):2204--2215, 2017.

\bibitem{DBLP:journals/ijcv/BegMTY05}
M.~F. Beg, M.~I. Miller, A.~Trouv{\'{e}}, and L.~Younes.
\newblock Computing large deformation metric mappings via geodesic flows of
  diffeomorphisms.
\newblock {\em International Journal of Computer Vision}, 61(2):139--157, 2005.

\bibitem{Bello2019}
G.~Bello, T.~Dawes, Jinming Duan, C.~Biffi, A.~de~Marvao, L.~Howard, S.~Gibbs,
  M.~Wilkins, S.~Cook, D.~Rueckert, and D.~O'Regan.
\newblock Deep learning cardiac motion analysis for human survival prediction.
\newblock {\em Nature Machine Intelligence}, 2019.

\bibitem{Bien2018}
N.~Bien, P.~Rajpurkar, R.~L. Ball, J.~Irvin, A.~Park, E.~Jones, M.~Bereket,
  B.~N. Patel, K.~W. Yeom, K.~Shpanskaya, S.~Halabi, E.~Zucker, G.~Fanton,
  D.~F. Amanatullah, C.~F. Beaulieu, G.~M. Riley, R.~J. Stewart, F.~G.
  Blankenberg, D.~B. Larson, R.~H. Jones, C.~P. Langlotz, A.~Y. Ng, and M.~P.
  Lungren.
\newblock Deep-learning-assisted diagnosis for knee magnetic resonance imaging:
  Development and retrospective validation of mrnet.
\newblock {\em PLOS Medicine}, 15(11):1--19, 2018.

\bibitem{DBLP:journals/corr/abs-1902-01046}
K.~Bonawitz, H.~Eichner, W.~Grieskamp, D.~Huba, A.Ingerman, V.~Ivanov,
  C.~Kiddon, J.~Konecn{\'{y}}, S.~Mazzocchi, H.~B. McMahan, T.~Van Overveldt,
  D.~Petrou, D.~Ramage, and J.~Roselander.
\newblock Towards federated learning at scale: System design.
\newblock {\em CoRR}, 2019.

\bibitem{Bookstein1991}
F.~Bookstein.
\newblock {\em Morphometric tools for landmark data}.
\newblock Cambridge University Press, 1991.

\bibitem{DBLP:conf/cvpr/BronsteinBMP10}
M.~M. Bronstein, A.~M. Bronstein, F.~Michel, and N.~Paragios.
\newblock Data fusion through cross-modality metric learning using
  similarity-sensitive hashing.
\newblock In {\em Computer Vision and Pattern Recognition (CVPR)}, pages
  3594--3601, 2010.

\bibitem{Ronneberger2016}
{\"{O}}.~{\c{C}}i{\c{c}}ek, A.~Abdulkadir, S.~S. Lienkamp, T.~Brox, and
  O.~Ronneberger.
\newblock 3d u-net: Learning dense volumetric segmentation from sparse
  annotation.
\newblock In {\em Medical Image Computing and Computer-Assisted Intervention
  (MICCAI)}, pages 424--432, 2016.

\bibitem{DBLP:conf/ipmi/CloughOBSK19}
J.~R. Clough, I.~{\"{O}}ks{\"{u}}z, N.~Byrne, J.~A. Schnabel, and A.~P. King.
\newblock Explicit topological priors for deep-learning based image
  segmentation using persistent homology.
\newblock In {\em Information Processing in Medical Imaging (IPMI)}, pages
  16--28, 2019.

\bibitem{GNC2014}
German National Cohort~(GNC) Consortium.
\newblock The {German National Cohort}: aims, study design and organization.
\newblock {\em European Journal of Epidemiology}, 29:371--382, 2014.

\bibitem{Cootes98}
T.~F. Cootes, G.~J. Edwards, and C.~J. Taylor.
\newblock Active appearance models.
\newblock In {\em European Conference on Computer Vision (ECCV)}, pages
  484--498, 1998.

\bibitem{Cootes95}
T.~F. Cootes, C.~J. Taylor, D.~H. Cooper, and J.~Graham.
\newblock Active shape models-their training and application.
\newblock {\em Computer Vision and Image Understanding}, 61(1):38--59, 1995.

\bibitem{Cortes1995}
C.~Cortes and V.~Vapnik.
\newblock Support-vector networks.
\newblock {\em Machine Learning}, 20(3):273--297, 1995.

\bibitem{Criminisi2012}
A.~Criminisi, J.~Shotton, and E.~Konukoglu.
\newblock Decision forests: {A} unified framework for classification,
  regression, density estimation, manifold learning and semi-supervised
  learning.
\newblock {\em Foundations and Trends in Computer Graphics and Vision},
  7(2-3):81--227, 2012.

\bibitem{DBLP:conf/miccai/DalcaBGS18}
A.~V. Dalca, G.~Balakrishnan, J.~V. Guttag, and M.~R. Sabuncu.
\newblock Unsupervised learning for fast probabilistic diffeomorphic
  registration.
\newblock In {\em Medical Image Computing and Computer Assisted Intervention
  (MICCAI)}, pages 729--738, 2018.

\bibitem{DBLP:journals/mia/VosBVSSI19}
B.~D. de~Vos, F.~F. Berendsen, M.~A. Viergever, H.~Sokooti, M.~Staring, and
  I.~Isgum.
\newblock A deep learning framework for unsupervised affine and deformable
  image registration.
\newblock {\em Medical Image Analysis}, 52:128--143, 2019.

\bibitem{Vos2017}
B.~D. de~Vos, J.~M. Wolterink, P.~A. de~Jong, T.~Leiner, M.~A. Viergever, and
  I.~Isgum.
\newblock Convnet-based localization of anatomical structures in 3-d medical
  images.
\newblock {\em {IEEE} Trans. Med. Imaging}, 36(7):1470--1481, 2017.

\bibitem{Donoho2006}
D.~L. Donoho.
\newblock Compressed sensing.
\newblock {\em {IEEE} Trans. Information Theory}, 52(4):1289--1306, 2006.

\bibitem{DBLP:conf/iccv/DosovitskiyFIHH15}
A.~Dosovitskiy, P.~Fischer, E.~Ilg, P.~H{\"{a}}usser, C.~Hazirbas, V.~Golkov,
  P.~van~der Smagt, D.~Cremers, and T.~Brox.
\newblock Flownet: Learning optical flow with convolutional networks.
\newblock In {\em International Conference on Computer Vision (ICCV)}, pages
  2758--2766, 2015.

\bibitem{Dwork:2012:FTA:2090236.2090255}
C.~Dwork, M.~Hardt, T.~Pitassi, O.~Reingold, and R.~Zemel.
\newblock Fairness through awareness.
\newblock In {\em Innovations in Theoretical Computer Science (ITCS)}, pages
  214--226, 2012.

\bibitem{Esteva2017}
A.~Esteva, B.~Kuprel, R.~Novoa, J.~Ko, S.~M. Swetter, H.~M. Blau, and S.~Thrun.
\newblock Dermatologist-level classification of skin cancer with deep neural
  networks.
\newblock {\em Nature}, 542, 2017.

\bibitem{DBLP:conf/miccai/FanCXYS18}
J.~Fan, X.~Cao, Z.~Xue, P.{-}T. Yap, and D.~Shen.
\newblock Adversarial similarity network for evaluating image alignment in deep
  learning based registration.
\newblock In {\em Medical Image Computing and Computer Assisted Intervention
  (MICCAI)}, pages 739--746, 2018.

\bibitem{Fauw2018}
J.~De Fauw, J.~R. Ledsam, B.~Romera-Paredes, S.~Nikolov, N.~Tomasev,
  S.Blackwell, H.~Askham, X.~Glorot, B.~O'Donoghue, D.~Visentin, G.~van~den
  Driessche, B.~Lakshminarayanan, C.~Meyer, F.~Mackinder, S.~Bouton, K.~Ayoub,
  R.~Chopra, D.~King, A.Karthikesalingam, C.~O Hughes, R.~Raine, J.~Hughes,
  D.~A. Sim, C.~Egan, A.~Tufail, H.~Montgomery, D.~Hassabis, G.~Rees, T.~Back,
  P.~T. Khaw, M.~Suleyman, J.~Cornebise, P.~A. Keane, and O.~Ronneberger.
\newblock Clinically applicable deep learning for diagnosis and referral in
  retinal disease.
\newblock {\em Nature Medicine}, 24:1342--1350, 2018.

\bibitem{10.1007/978-3-319-66179-7_65}
X.~Feng, J.~Yang, A.~F. Laine, and E.~D. Angelini.
\newblock Discriminative localization in cnns for weakly-supervised
  segmentation of pulmonary nodules.
\newblock In {\em Medical Image Computing and Computer Assisted Intervention
  ('MICCAI)}, pages 568--576, 2017.

\bibitem{Finlayson2019}
S.~G. Finlayson, J.~D. Bowers, J.~Ito, J.~L. Zittrain, A.~L. Beam, and I.~S.
  Kohane.
\newblock Adversarial attacks on medical machine learning.
\newblock {\em Science}, 363(6433):1287--1289, 2019.

\bibitem{GASER20011140}
C.~Gaser, I.~Nenadic, B.~R. Buchsbaum, E.~A. Hazlett, and M.~S. Buchsbaum.
\newblock Deformation-based morphometry and its relation to conventional
  volumetry of brain lateral ventricles in {MRI}.
\newblock {\em NeuroImage}, 13(6):1140 -- 1145, 2001.

\bibitem{Ghafoorian2017}
M.~Ghafoorian, A.~Mehrtash, T.~Kapur, N.~Karssemeijer, E.~Marchiori,
  M.~Pesteie, C.~R.~G. Guttmann, F.{-}E. de~Leeuw, C.~M. Tempany, B.~van
  Ginneken, A.~Fedorov, P.~Abolmaesumi, B.~Platel, and W.~M.~Wells III.
\newblock Transfer learning for domain adaptation in {MRI:} application in
  brain lesion segmentation.
\newblock In {\em Medical Image Computing and Computer Assisted Intervention
  (MICCAI)}, pages 516--524, 2017.

\bibitem{DBLP:journals/corr/abs-1906-07775}
F.~C. Ghesu, B.~Georgescu, E.~Gibson, S.~G{\"{u}}ndel, M.~K. Kalra, R.~Singh,
  S.~R. Digumarthy, S.~Grbic, and D.~Comaniciu.
\newblock Quantifying and leveraging classification uncertainty for chest
  radiograph assessment.
\newblock {\em CoRR}, abs/1906.07775, 2019.

\bibitem{pmlr-v48-gilad-bachrach16}
R.~Gilad-Bachrach, N.~Dowlin, K.~Laine, K.~Lauter, M.~Naehrig, and J.~Wernsing.
\newblock Cryptonets: Applying neural networks to encrypted data with high
  throughput and accuracy.
\newblock In {\em International Conference on Machine Learning (ICML)},
  volume~48, pages 201--210, 2016.

\bibitem{DBLP:conf/nips/GoodfellowPMXWOCB14}
I.~J. Goodfellow, J.~Pouget{-}Abadie, M.~Mirza, B.~Xu, D.~Warde{-}Farley,
  S.~Ozair, A.~C. Courville, and Y.~Bengio.
\newblock Generative adversarial nets.
\newblock In {\em Advances in Neural Information Processing Systems (NIPS)},
  pages 2672--2680, 2014.

\bibitem{Goodfellow2015}
I.~J. Goodfellow, J.~Shlens, and C.~Szegedy.
\newblock Explaining and harnessing adversarial examples.
\newblock In {\em International Conference on Learning Representations (ICLR)},
  2015.

\bibitem{DBLP:journals/corr/Grgic-HlacaZGW17}
N.~Grgic{-}Hlaca, M.~B. Zafar, K.~P. Gummadi, and A.~Weller.
\newblock On fairness, diversity and randomness in algorithmic decision making.
\newblock {\em CoRR}, 2017.

\bibitem{Gunning2019}
D.~Gunning.
\newblock Darpa's explainable artificial intelligence {(XAI)} program.
\newblock In {\em 24th International Conference on Intelligent User Interfaces,
  (IUI)}, 2019.

\bibitem{Hammernik2018}
K.~Hammernik, T.~Klatzer, E.~Kobler, M.~P. Recht, D.~K. Sodickson, T.~Pock, and
  F.~Knoll.
\newblock Learning a variational network for reconstruction of accelerated mri
  data.
\newblock {\em Magnetic Resonance in Medicine}, 79(6):3055--3071, 2018.

\bibitem{Havaei2017}
M.~Havaei, A.~Davy, D.~Warde{-}Farley, A.~Biard, A.~C. Courville, Y.~Bengio,
  C.~Pal, P.{-}M. Jodoin, and H.~Larochelle.
\newblock Brain tumor segmentation with deep neural networks.
\newblock {\em Medical Image Analysis}, 35:18--31, 2017.

\bibitem{Heckemann2006}
R.~A. Heckemann, J.~V. Hajnal, P.~Aljabar, D.~Rueckert, and A.~Hammers.
\newblock Automatic anatomical brain {MRI} segmentation combining label
  propagation and decision fusion.
\newblock {\em NeuroImage}, 33(1):115--126, 2006.

\bibitem{DBLP:journals/corr/abs-1812-05720}
M.~Hein, M.~Andriushchenko, and J.~Bitterwolf.
\newblock Why relu networks yield high-confidence predictions far away from the
  training data and how to mitigate the problem.
\newblock In {\em Computer Vision and Pattern Recognition (CVPR)}, 2019.

\bibitem{DBLP:journals/mia/HeinrichJBMGBS12}
M.~P. Heinrich, M.~Jenkinson, M.~Bhushan, T.~N. Matin, F.~Gleeson, M.~Brady,
  and J.~A. Schnabel.
\newblock {MIND:} modality independent neighbourhood descriptor for multi-modal
  deformable registration.
\newblock {\em Medical Image Analysis}, 16(7):1423--1435, 2012.

\bibitem{pmlr-v80-hoffman18a}
J.~Hoffman, E.~Tzeng, T.~Park, J.-Y. Zhu, P.~Isola, K.~Saenko, A.~Efros, and
  T.~Darrell.
\newblock {C}y{CADA}: Cycle-consistent adversarial domain adaptation.
\newblock In {\em International Conference on Machine Learning (ICML)}, pages
  1989--1998, 2018.

\bibitem{DBLP:journals/ai/HornS81}
B.~K.~P. Horn and B.~G. Schunck.
\newblock Determining optical flow.
\newblock {\em Artif. Intell.}, 17(1-3):185--203, 1981.

\bibitem{HUA2008458}
X.~Hua, A.~D. Leow, N.~Parikshak, S.~Lee, M.-C. Chiang, A.~W. Toga, C.~R. Jack,
  M.~W. Weiner, and P.~M. Thompson.
\newblock Tensor-based morphometry as a neuroimaging biomarker for alzheimer's
  disease: An mri study of 676 ad, mci, and normal subjects.
\newblock {\em NeuroImage}, 43(3):458 -- 469, 2008.

\bibitem{DBLP:journals/neuroimage/HuizingaPVRBIRN18}
W.~Huizinga, D.~H.~J. Poot, M.~W. Vernooij, G.~Roshchupkin, E.~Bron, M.~A.
  Ikram, D.~Rueckert, W.~J. Niessen, and S.~Klein.
\newblock A spatio-temporal reference model of the aging brain.
\newblock {\em NeuroImage}, 169:11--22, 2018.

\bibitem{Irvin2019}
J.~Irvin, P.~Rajpurkar, M.~Ko, Y.~Yu, S.~Ciurea{-}Ilcus, C.~Chute, H.~Marklund,
  B.~Haghgoo, R.~L. Ball, K.~Shpanskaya, J.~Seekins, D.~A. Mong, S.~S. Halabi,
  J.~K. Sandberg, R.~Jones, D.~B. Larson, C.~P. Langlotz, B.~N. Patel, M.~P.
  Lungren, and A.~Y. Ng.
\newblock Chexpert: {A} large chest radiograph dataset with uncertainty labels
  and expert comparison.
\newblock {\em CoRR}, abs/1901.07031, 2019.

\bibitem{DBLP:conf/nips/JaderbergSZK15}
M.~Jaderberg, K.~Simonyan, A.~Zisserman, and K.~Kavukcuoglu.
\newblock Spatial transformer networks.
\newblock In {\em Advances in Neural Information Processing Systems (NIPS)},
  pages 2017--2025, 2015.

\bibitem{Yan2018}
L.~Lu K.~Yan, X.~Wang and R.~M. Summers.
\newblock Deeplesion: automated mining of large-scale lesion annotations and
  universal lesion detection with deep learning.
\newblock {\em Journal of Medical Imaging}, 5(3):1 -- 11, 2018.

\bibitem{Kamnitsas2017a}
K.~Kamnitsas, C.~F. Baumgartner, C.~Ledig, V.~F.~J. Newcombe, J.~P. Simpson,
  A.~D. Kane, D.~K. Menon, A.~V. Nori, A.~Criminisi, D.~Rueckert, and
  B.~Glocker.
\newblock Unsupervised domain adaptation in brain lesion segmentation with
  adversarial networks.
\newblock In {\em Information Processing in Medical Imaging (IPMI)}, pages
  597--609, 2017.

\bibitem{Kamnitsas2017}
K.~Kamnitsas, C.~Ledig, V.~F.~J. Newcombe, J.~P. Simpson, Andrew~D. Kane, D.~K.
  Menon, D.~Rueckert, and B.~Glocker.
\newblock Efficient multi-scale 3d {CNN} with fully connected {CRF} for
  accurate brain lesion segmentation.
\newblock {\em Medical Image Analysis}, 36:61--78, 2017.

\bibitem{Katz2017}
G.~Katz, C.~Barrett, D.~L. Dill, K.~Julian, and M.~J. Kochenderfer.
\newblock Towards proving the adversarial robustness of deep neural networks.
\newblock In {\em Workshop on Formal Verification of Autonomous Vehicles},
  pages 19--26, 2017.

\bibitem{DBLP:journals/neuroimage/KleinGAYFAGMP10}
A.~Klein, S.~S. Ghosh, B.~B. Avants, B.~T.~T. Yeo, B.~Fischl, B.~A. Ardekani,
  J.~C. Gee, J.~J. Mann, and R.~V. Parsey.
\newblock Evaluation of volume-based and surface-based brain image registration
  methods.
\newblock {\em NeuroImage}, 51(1):214--220, 2010.

\bibitem{Kloeppel2008}
S.~Kl\"{o}ppel, C.~M. Stonnington, C.~Chu, B.~Draganski, R.~I. Scahill, J.~D.
  Rohrer, N.~C. Fox, Jr~Jack, C.~R., J.~Ashburner, and R.~S.~J. Frackowiak.
\newblock {Automatic classification of MR scans in Alzheimer's disease}.
\newblock {\em Brain}, 131(3):681--689, 2008.

\bibitem{Kohavi1995}
R.~Kohavi.
\newblock A study of cross-validation and bootstrap for accuracy estimation and
  model selection.
\newblock In {\em International Joint Conference on Artificial Intelligence
  (IJCAI)}, pages 1137--1145, 1995.

\bibitem{Koikkalainen2011}
J.~Koikkalainen, J.~Lotjonen, L.~Thurfjell, D.~Rueckert, G.~Waldemar, and
  H.~Soininen.
\newblock Multi-template tensor-based morphometry: Application to analysis of
  alzheimer's disease.
\newblock {\em NeuroImage}, 56(3):1134 -- 1144, 2011.

\bibitem{DBLP:journals/corr/abs-1907-13524}
J.~{Krebs}, H.~{Delingette}, B.~{Mailhé}, N.~{Ayache}, and T.~{Mansi}.
\newblock Learning a probabilistic model for diffeomorphic registration.
\newblock {\em IEEE Transactions on Medical Imaging}, 38(9):2165--2176, 2019.

\bibitem{LeCun1998}
Y.~{Lecun}, L.~{Bottou}, Y.~{Bengio}, and P.~{Haffner}.
\newblock Gradient-based learning applied to document recognition.
\newblock {\em Proceedings of the IEEE}, 86(11):2278--2324, 1998.

\bibitem{8672808}
M.~C.~H. {Lee}, K.~{Petersen}, N.~{Pawlowski}, B.~{Glocker}, and M.~{Schaap}.
\newblock Tetris: Template transformer networks for image segmentation with
  shape priors.
\newblock {\em IEEE Transactions on Medical Imaging}, page in press, 2019.

\bibitem{Leemput99}
K.~Van Leemput, F.~Maes, D.~Vandermeulen, and P.~Suetens.
\newblock Automated model-based tissue classification of {MR} images of the
  brain.
\newblock {\em {IEEE} Trans. Med. Imaging}, 18(10):897--908, 1999.

\bibitem{Leibig2017}
C.~Leibig, V.~Allken, M.~S. Ayhan, P.~Berens, and S.~Wahl.
\newblock Leveraging uncertainty information from deep neural networks for
  disease detection.
\newblock {\em Scientific Reports}, 7, 2017.

\bibitem{Leofante2018}
F.~Leofante, N.~Narodytska, L.~Pulina, and A.~Tacchella.
\newblock Automated verification of neural networks: Advances, challenges and
  perspectives.
\newblock {\em CoRR}, abs/1805.09938, 2018.

\bibitem{10.1007/978-3-319-10602-1_48}
T.-Y. Lin, M.~Maire, S.~Belongie, J.~Hays, P.~Perona, D.~Ramanan,
  P.~Doll{\'a}r, and C.~Lawrence Zitnick.
\newblock Microsoft coco: Common objects in context.
\newblock In {\em European Conference on Computer Vision (ECCV)}, pages
  740--755, 2014.

\bibitem{Lindner2013}
C.~Lindner, S.~Thiagarajah, J.M. Wilkinson, G.A. Wallis, and T.F. Cootes.
\newblock Development of a fully automatic shape model matching (fasmm) system
  to derive statistical shape models from radiographs: application to the
  accurate capture and global representation of proximal femur shape.
\newblock {\em Osteoarthritis and Cartilage}, 21(10):1537 -- 1544, 2013.

\bibitem{DBLP:journals/ijcv/LombaertGPAC14}
H.~Lombaert, L.~Grady, X.~Pennec, N.~Ayache, and F.~Cheriet.
\newblock Spectral log-demons: Diffeomorphic image registration with very large
  deformations.
\newblock {\em International Journal of Computer Vision}, 107(3):254--271,
  2014.

\bibitem{Long2015}
J.~Long, E.~Shelhamer, and T.~Darrell.
\newblock Fully convolutional networks for semantic segmentation.
\newblock In {\em Computer Vision and Pattern Recognition (CVPR)}, pages
  3431--3440, 2015.

\bibitem{DBLP:journals/neuroimage/LorenziAFP13}
M.~Lorenzi, N.~Ayache, G.~B. Frisoni, and X.~Pennec.
\newblock Lcc-demons: {A} robust and accurate symmetric diffeomorphic
  registration algorithm.
\newblock {\em NeuroImage}, 81:470--483, 2013.

\bibitem{Lustig2007}
M.~Lustig, D.~Donoho, and J.~M. Pauly.
\newblock Sparse {MRI}: The application of compressed sensing for rapid {MR}
  imaging.
\newblock {\em Magnetic Resonance in Medicine}, 58(6):1182--1195, 2007.

\bibitem{DBLP:journals/tmi/MaesCVMS97}
F.~Maes, A.~Collignon, D.~Vandermeulen, G.~Marchal, and P.~Suetens.
\newblock Multimodality image registration by maximization of mutual
  information.
\newblock {\em {IEEE} Trans. Med. Imaging}, 16(2):187--198, 1997.

\bibitem{Mahendran2015}
A.~Mahendran and A.~Vedaldi.
\newblock Understanding deep image representations by inverting them.
\newblock In {\em Computer Vision and Pattern Recognition (CVPR)}, pages
  5188--5196, 2015.

\bibitem{DBLP:journals/ijcv/MansiPSDA11}
T.~Mansi, X.~Pennec, M.~Sermesant, H.~Delingette, and N.~Ayache.
\newblock ilogdemons: {A} demons-based registration algorithm for tracking
  incompressible elastic biological tissues.
\newblock {\em International Journal of Computer Vision}, 92(1):92--111, 2011.

\bibitem{McInerney1996}
T.~McInerney and D.~Terzopoulos.
\newblock Deformable models in medical image analysis: a survey.
\newblock {\em Medical Image Analysis}, 1(2):91--108, 1996.

\bibitem{Miller2016}
K.~Miller, F.~Alfaro-Almagro, N.~Bangerter, D.~Thomas, E.~Yacoub, J.~Xu,
  A.~Bartsch, S.~Jbabdi, S.~Sotiropoulos, J.~Andersson, L.~Griffanti,
  G.~Douaud, T.~Okell, P.~Weale, I.~Dragonu, S.~Garratt, S.~Hudson, R.~Collins,
  M.~Jenkinson, and S.~Smith.
\newblock Multimodal population brain imaging in the {UK Biobank} prospective
  epidemiological study.
\newblock {\em Nature Meuroscience}, 19, 2016.

\bibitem{Montillo2011}
A.~Montillo, J.~Shotton, J.~M. Winn, J.~E. Iglesias, D.~N. Metaxas, and
  A.~Criminisi.
\newblock Entangled decision forests and their application for semantic
  segmentation of {CT} images.
\newblock In {\em Information Processing in Medical Imaging (IPMI)}, pages
  184--196, 2011.

\bibitem{Oksuz2018}
I.~{\"{O}}ks{\"{u}}z, B.~Ruijsink, E.~Puyol{-}Ant{\'{o}}n, A.~Bustin, G.~Cruz,
  C.~Prieto, D.~Rueckert, J.~A. Schnabel, and A.~P. King.
\newblock Deep learning using k-space based data augmentation for automated
  cardiac {MR} motion artefact detection.
\newblock In {\em Medical Image Computing and Computer Assisted Intervention
  (MICCAI)}, pages 250--258, 2018.

\bibitem{Oktay2018}
O.~Oktay, E.~Ferrante, K.~Kamnitsas, M.~P. Heinrich, W.~Bai, J.~Caballero,
  S.~A. Cook, A.~de~Marvao, T.~Dawes, D.~P. O'Regan, B.~Kainz, B.~Glocker, and
  D.~Rueckert.
\newblock Anatomically constrained neural networks ({ACNNs}): Application to
  cardiac image enhancement and segmentation.
\newblock {\em {IEEE} Trans. Med. Imaging}, 37(2):384--395, 2018.

\bibitem{DBLP:journals/mia/PapiezHFRS14}
B.~W. Papiez, M.~P. Heinrich, J.~Fehrenbach, L.~Risser, and J.~A. Schnabel.
\newblock An implicit sliding-motion preserving regularisation via bilateral
  filtering for deformable image registration.
\newblock {\em Medical Image Analysis}, 18(8):1299--1311, 2014.

\bibitem{Petersen2013}
S.~Petersen, P.~Matthews, F.~Bamberg, D.~Bluemke, J.~Francis, M.~Friedrich,
  P.~Leeson, E.~Nagel, S.~Plein, F.~Rademakers, A.~Young, S.~Garratt,
  T.~Peakman, J.~Sellors, R.~Collins, and S/~Neubauer.
\newblock Imaging in population science: Cardiovascular magnetic resonance in
  100,000 participants of {UK Biobank} - rationale, challenges and approaches.
\newblock {\em Journal of Cardiovascular Magnetic Resonance}, 15:46, 2013.

\bibitem{DBLP:conf/miccai/QinBSPPNR18}
C.~Qin, W.~Bai, J.~Schlemper, S.~E. Petersen, S.~K. Piechnik, S.~Neubauer, and
  D.~Rueckert.
\newblock Joint learning of motion estimation and segmentation for cardiac {MR}
  image sequences.
\newblock In {\em Medical Image Computing and Computer Assisted Intervention
  (MICCAI)}, pages 472--480, 2018.

\bibitem{DBLP:journals/tmi/RajchlLOKPBDRHK17}
M.~Rajchl, M.~C.~H. Lee, O.~Oktay, K.~Kamnitsas, J.~Passerat{-}Palmbach,
  W.~Bai, M.~Damodaram, M.~A. Rutherford, J.~V. Hajnal, B.~Kainz, and
  D.~Rueckert.
\newblock Deepcut: Object segmentation from bounding box annotations using
  convolutional neural networks.
\newblock {\em {IEEE} Transactions on Medical Imaging}, 36(2):674--683, 2017.

\bibitem{Rajpurkar2018}
P.~Rajpurkar, J.~Irvin, R.~L. Ball, K.~Zhu, B.~Yang, H.~Mehta, T.~Duan,
  D.~Ding, A.~Bagul, C.~P. Langlotz, B.~N. Patel, K.~W. Yeom, K.~Shpanskaya,
  F.~G. Blankenberg, J.~Seekins, T.~J. Amrhein, D.~A. Mong, S~S. Halabi, E.~J.
  Zucker, A.~Y. Ng, and M.P. Lungren.
\newblock Deep learning for chest radiograph diagnosis: A retrospective
  comparison of the chexnext algorithm to practicing radiologists.
\newblock {\em PLoS Medicine}, 15(11):1--17, 2018.

\bibitem{DBLP:conf/cvpr/RanjanB17}
A.~Ranjan and M.~J. Black.
\newblock Optical flow estimation using a spatial pyramid network.
\newblock In {\em Computer Vision and Pattern Recognition (CVPR) 2017}, pages
  2720--2729, 2017.

\bibitem{Ravishankar2011}
S.~Ravishankar and Y.~Bresler.
\newblock {MR} image reconstruction from highly undersampled k-space data by
  dictionary learning.
\newblock {\em {IEEE} Trans. Med. Imaging}, 30(5):1028--1041, 2011.

\bibitem{Ronneberger2015}
O.~Ronneberger, P.~Fischer, and T.~Brox.
\newblock U-net: Convolutional networks for biomedical image segmentation.
\newblock In {\em Medical Image Computing and Computer-Assisted Intervention
  (MICCAI)}, pages 234--241, 2015.

\bibitem{DBLP:journals/tmi/RueckertSHHLH99}
D.~Rueckert, L.~I. Sonoda, C.~Hayes, D.~L.~G. Hill, M.~O. Leach, and D.~J.
  Hawkes.
\newblock Non-rigid registration using free-form deformations: Application to
  breast {MR} images.
\newblock {\em {IEEE} Trans. Med. Imaging}, 18(8):712--721, 1999.

\bibitem{ILSVRC15}
O.~Russakovsky, J.~Deng, H.~Su, J.~Krause, S.~Satheesh, S.~Ma, Z.~Huang,
  A.~Karpathy, A.~Khosla, M.~Bernstein, A.~C. Berg, and L.~Fei-Fei.
\newblock {ImageNet Large Scale Visual Recognition Challenge}.
\newblock {\em International Journal of Computer Vision}, 115(3):211--252,
  2015.

\bibitem{6582713}
A.~D. {Sarwate} and K.~{Chaudhuri}.
\newblock Signal processing and machine learning with differential privacy:
  Algorithms and challenges for continuous data.
\newblock {\em IEEE Signal Processing Magazine}, 30(5):86--94, 2013.

\bibitem{Schlemper2018}
J.~Schlemper, J.~Caballero, J.~V. Hajnal, A.~N. Price, and D.~Rueckert.
\newblock A deep cascade of convolutional neural networks for dynamic {MR}
  image reconstruction.
\newblock {\em {IEEE} Trans. Med. Imaging}, 37(2):491--503, 2018.

\bibitem{Schlemper2019}
J.~Schlemper, O.~Oktay, M.~Schaap, M.~P. Heinrich, B.~Kainz, B.~Glocker, and
  D.~Rueckert.
\newblock Attention gated networks: Learning to leverage salient regions in
  medical images.
\newblock {\em Medical Image Analysis}, 53:197--207, 2019.

\bibitem{10.1007/978-3-030-11723-8_9}
M.~J. Sheller, G.~A. Reina, B.~Edwards, J.~Martin, and S.~Bakas.
\newblock Multi-institutional deep learning modeling without sharing patient
  data: A feasibility study on brain tumor segmentation.
\newblock In {\em Brainlesion: Glioma, Multiple Sclerosis, Stroke and Traumatic
  Brain Injuries}, pages 92--104, 2019.

\bibitem{Shin2016}
H.{-}C. Shin, H.~R. Roth, M.~Gao, L.~Lu, Z.~Xu, I.~Nogues, J.~Yao, D.~J.
  Mollura, and R.~M. Summers.
\newblock Deep convolutional neural networks for computer-aided detection:
  {CNN} architectures, dataset characteristics and transfer learning.
\newblock {\em {IEEE} Trans. Med. Imaging}, 35(5):1285--1298, 2016.

\bibitem{Shrikumar2017}
A.~Shrikumar, P.~Greenside, and A.~Kundaje.
\newblock Learning important features through propagating activation
  differences.
\newblock In {\em International Conference on Machine Learning (ICML)}, pages
  3145--3153, 2017.

\bibitem{Simonyan2014a}
K.~Simonyan, A.~Vedaldi, and A.~Zisserman.
\newblock Deep inside convolutional networks: Visualising image classification
  models and saliency maps.
\newblock In {\em International Conference on Learning Representations (ICLR)},
  2014.

\bibitem{Simonyan2014}
K.~Simonyan and A.~Zisserman.
\newblock Very deep convolutional networks for large-scale image recognition.
\newblock In {\em International Conference on Learning Representations (ICLR)},
  2015.

\bibitem{Springenberg2015}
J.~T. Springenberg, A.~Dosovitskiy, T.~Brox, and M.~A. Riedmiller.
\newblock Striving for simplicity: The all convolutional net.
\newblock In {\em International Conference on Learning Representations (ICLR)},
  2015.

\bibitem{DBLP:journals/pr/StudholmeHH99}
C.~Studholme, D.~L.~G. Hill, and D.~J. Hawkes.
\newblock An overlap invariant entropy measure of 3d medical image alignment.
\newblock {\em Pattern Recognition}, 32(1):71--86, 1999.

\bibitem{Styner2004}
M.~Styner, J.~A. Lieberman, D.~Pantazis, and G.~Gerig.
\newblock Boundary and medial shape analysis of the hippocampus in
  schizophrenia.
\newblock {\em Medical Image Analysis}, 8(3):197 -- 203, 2004.
\newblock Medical Image Computing and Computer-Assisted Intervention - MICCAI
  2003.

\bibitem{Topol2014}
E.~Topol.
\newblock Individualized medicine from prewomb to tomb.
\newblock {\em Cell}, 157:241--253, 2014.

\bibitem{Tran2016}
P.~V. Tran.
\newblock A fully convolutional neural network for cardiac segmentation in
  short-axis {MRI}.
\newblock {\em CoRR}, abs/1604.00494, 2016.

\bibitem{DBLP:journals/neuroimage/VercauterenPPA09}
T.~Vercauteren, X.~Pennec, A.~Perchant, and N.~Ayache.
\newblock Diffeomorphic demons: Efficient non-parametric image registration.
\newblock {\em NeuroImage}, 45(1):S61--S72, 2009.

\bibitem{VROOMAN200771}
H.~A. Vrooman, C.~A. Cocosco, F.~van~der Lijn, R.~Stokking, M.~A. Ikram, M.~W.
  Vernooij, M.~M.B. Breteler, and W.~J. Niessen.
\newblock Multi-spectral brain tissue segmentation using automatically trained
  k-nearest-neighbor classification.
\newblock {\em NeuroImage}, 37(1):71 -- 81, 2007.

\bibitem{Wang2013}
H.~Wang, J.~W. Suh, S.~R. Das, J.~Pluta, C.~Craige, and P.~A. Yushkevich.
\newblock Multi-atlas segmentation with joint label fusion.
\newblock {\em {IEEE} Trans. Pattern Anal. Mach. Intell.}, 35(3):611--623,
  2013.

\bibitem{Wang2016}
S.~Wang, Z.~Su, L.~Ying, X.~Peng, S.~Zhu, F.~Liang, D.~Feng, and D.~Liang.
\newblock Accelerating magnetic resonance imaging via deep learning.
\newblock In {\em {IEEE} International Symposium on Biomedical Imaging (ISBI)},
  pages 514--517, 2016.

\bibitem{WARFIELD200043}
S.~K. Warfield, M.~Kaus, F.~A. Jolesz, and R.~Kikinis.
\newblock Adaptive, template moderated, spatially varying statistical
  classification.
\newblock {\em Medical Image Analysis}, 4(1):43 -- 55, 2000.

\bibitem{DBLP:conf/iccv/WeinzaepfelRHS13}
P.~Weinzaepfel, J.~Revaud, Z.~Harchaoui, and C.~Schmid.
\newblock Deepflow: Large displacement optical flow with deep matching.
\newblock In {\em International Conference on Computer Vision (ICCV)}, pages
  1385--1392, 2013.

\bibitem{DBLP:journals/corr/abs-1708-01870}
A.~Weller.
\newblock Challenges for transparency.
\newblock {\em CoRR}, 2017.

\bibitem{Wells96}
W.~M. {Wells III}, W.~E.~L. Grimson, R.~Kikinis, and F.~A. Jolesz.
\newblock Adaptive segmentation of {MRI} data.
\newblock {\em {IEEE} Trans. Med. Imaging}, 15(4):429--442, 1996.

\bibitem{DBLP:journals/mia/WellsVANK96}
W.~M. {Wells III}, P.~A. Viola, H.~Atsumi, S.~Nakajima, and R.~Kikinis.
\newblock Multi-modal volume registration by maximization of mutual
  information.
\newblock {\em Medical Image Analysis}, 1(1):35--51, 1996.

\bibitem{DBLP:conf/miccai/WrightKGSMHRS18}
R.~Wright, B.~Khanal, A.~G{\'{o}}mez, E.Skelton, J.~Matthew, J.~V. Hajnal,
  D.~Rueckert, and J.~A. Schnabel.
\newblock {LSTM} spatial co-transformer networks for registration of 3d fetal
  {US} and {MR} brain images.
\newblock In {\em Preterm, Perinatal and Paediatric Image Analysis Workshop
  (PIPPI)}, pages 149--159, 2018.

\bibitem{Wu2017}
L.~Wu, J.{-}Z. Cheng, S.~Li, B.~Y. Lei, T.~Wang, and D.~Ni.
\newblock {FUIQA:} fetal ultrasound image quality assessment with deep
  convolutional networks.
\newblock {\em {IEEE} Trans. Cybernetics}, 47(5):1336--1349, 2017.

\bibitem{DBLP:journals/neuroimage/YangKSN17}
X.~Yang, R.~Kwitt, M.~Styner, and M.~Niethammer.
\newblock Quicksilver: Fast predictive image registration - {A} deep learning
  approach.
\newblock {\em NeuroImage}, 158:378--396, 2017.

\bibitem{Zeiler2014}
M.~D. Zeiler and R.~Fergus.
\newblock Visualizing and understanding convolutional networks.
\newblock In {\em European Conference Computer Vision (ECCV)}, pages 818--833,
  2014.

\bibitem{6116655}
K.~{Zhang}, J.~{Deng}, and W.~{Lu}.
\newblock Segmenting human knee cartilage automatically from multi-contrast mr
  images using support vector machines and discriminative random fields.
\newblock In {\em International Conference on Image Processing (ICIP)}, pages
  721--724, 2011.

\bibitem{Zhu2017}
B.~Zhu, J.~Liu, B.~Rosen, and M.~Rosen.
\newblock Image reconstruction by domain transform manifold learning.
\newblock {\em Nature}, 555, 2017.

\bibitem{Zikic2012}
D.~Zikic, B.~Glocker, E.~Konukoglu, A.~Criminisi, {\c{C}}.~Demiralp,
  J.~Shotton, O.~M. Thomas, T.~Das, R.~Jena, and S.~J. Price.
\newblock Decision forests for tissue-specific segmentation of high-grade
  gliomas in multi-channel {MR}.
\newblock In {\em Medical Image Computing and Computer-Assisted Intervention
  (MICCAI)}, pages 369--376, 2012.

\end{thebibliography}

\end{document}